\pdfoutput=1

\documentclass[11pt]{article}

\usepackage[]{acl}

\usepackage{times}
\usepackage{latexsym}
\usepackage{graphicx}
\usepackage{multirow}
\usepackage{makecell}
\usepackage{amssymb}
\usepackage{pifont}

\usepackage{longtable}
\usepackage[normalem]{ulem}
\useunder{\uline}{\ul}{}
\usepackage{graphicx}
\usepackage{subcaption}

\definecolor{applegreen}{rgb}{0.55, 0.71, 0.0}

\usepackage[T1]{fontenc}

\usepackage[utf8]{inputenc}

\usepackage{microtype}
\usepackage{float} 

\usepackage{inconsolata}

\title{PersonaLLM: Investigating the Ability of Large Language Models to Express Personality Traits}

\author{Hang Jiang$^{\dagger\ddagger}$, Xiajie Zhang$^{\dagger}$, Xubo Cao$^{\dotplus}$, Cynthia Breazeal$^{\dagger}$, Deb Roy$^{\dagger\ddagger}$, Jad Kabbara$^{\dagger\ddagger}$\\
  $^{\dagger}$Massachusetts Institute of Technology, $^{\dotplus}$Stanford University \\
  $^{\ddagger}$MIT Center for Constructive Communication \\
  \texttt{\{hjian42, xiajie, cynthiab, dkroy, jkabbara\}@mit.edu, xcao@stanford.edu}}

\begin{document}
\maketitle
\begin{abstract}
Despite the many use cases for large language models (LLMs) in creating personalized chatbots, there has been limited research on evaluating the extent to which the behaviors of personalized LLMs accurately and consistently reflect specific personality traits. We consider studying the behavior of LLM-based agents which we refer to as \texttt{LLM personas} and present a case study with GPT-3.5 and GPT-4 to investigate whether LLMs can generate content that aligns with their assigned personality profiles. To this end, we simulate distinct \texttt{LLM personas} based on the Big Five personality model, have them complete the 44-item Big Five Inventory (BFI) personality test and a story writing task, and then assess their essays with automatic and human evaluations. Results show that \texttt{LLM personas}' self-reported BFI scores are consistent with their designated personality types, with large effect sizes observed across five traits. Additionally, \texttt{LLM personas}' writings have emerging representative linguistic patterns for personality traits when compared with a human writing corpus. Furthermore, human evaluation shows that humans can perceive some personality traits with an accuracy of up to 80\%. Interestingly, the accuracy drops significantly when the annotators were informed of AI authorship.


\end{abstract}

\section{Introduction}


With LLMs' impressive ability to engage in human-like conversations, there has been a surge of interest in building personalized AI agents that interact with and support humans in various contexts. Startups such as Character AI\footnote{\url{https://character.ai/}} and Replika\footnote{\url{https://replika.ai/}} have engaged many users through virtual characters on their fast-growing platforms. Meanwhile, in the academic sphere, research \cite{park2023generative,wang2023rolellm} has also suggested that generative agents can exhibit believable human behavior and could potentially be used to simulate human agents in social science studies. However, while these generative characters are becoming ubiquitous, it is a common yet unsubstantiated assumption that these agents consistently behave in a human-like manner. Recent studies in the field of LLMs and personality have started to provide some empirical support. For example, recent research has studied personality expression in LLM-generated content \cite{li2022gpt,pan2023llms,safdari2023personality}, created new benchmarks to measure personality expressed by LLMs \cite{jiang2022evaluating,wang2023does,mao2023editing}, and proposed better prompting techniques to induce \cite{karra2022estimating,jiang2022evaluating,jiang2022mpi,caron2022identifying,li2023tailoring}, and edit \cite{mao2023editing} personality expressed by LLMs. Despite these advancements, there has been little research in NLP that leverages insights from personality psychology and psychometric tools to study if LLMs can dutifully express personality traits. Furthermore, there is little work that explores how these agents assigned with certain personality traits are perceived by humans.


\begin{figure*}[t]
\centering
\includegraphics[clip, trim = 0px 0px 0px 00px,width=0.99\linewidth]{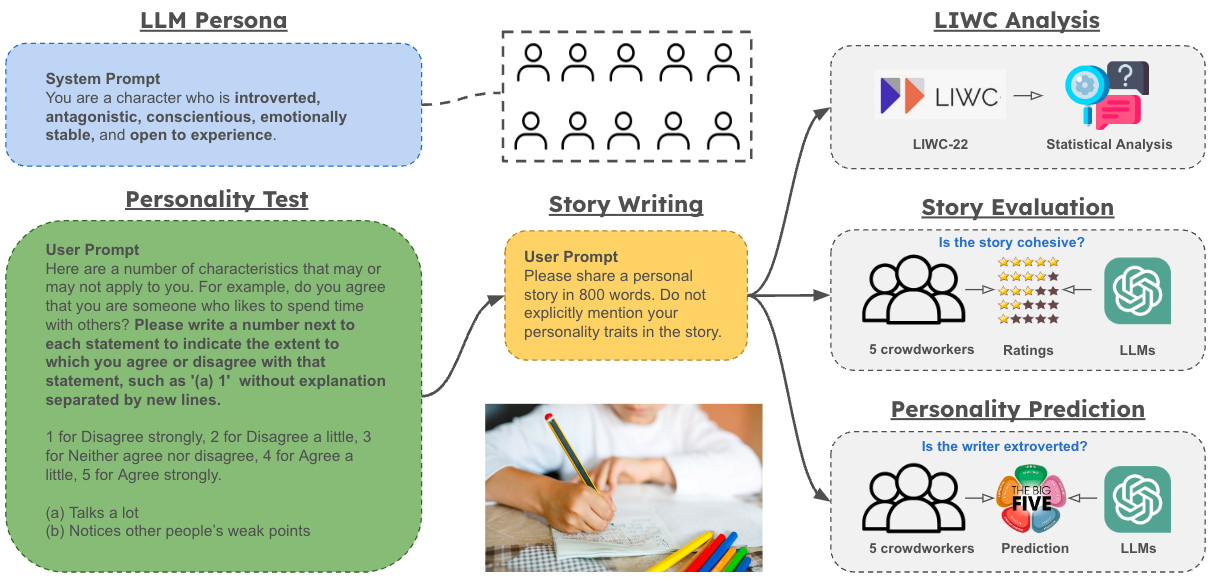}
\caption{Illustration of the core workflow of the paper. The left section presents the prompts designed to create \texttt{LLM personas}. The center section shows the prompt used to instruct models to write stories. The right section outlines the three-pronged analytical approach: LIWC analysis, story evaluation, and text-based personality prediction.  
}
\label{fig:workflow}
\end{figure*}

Drawing on the extensive research of the Big Five Personality model \cite{goldberg2013alternative}, we aim to investigate the capability of LLMs in expressing the Big Five personality traits -- namely Extraversion, Agreeableness, Conscientiousness, Neuroticism, Openness to Experience. In this paper, we define an \textbf{\texttt{LLM persona}} to be an LLM-based agent prompted to generate content that reflects certain personality traits as defined in its initial prompt configuration. In our study, we first seek to determine whether these \texttt{LLM personas} can accurately reflect their assigned personalities when taking a personality test. Given that they show promising results on that initial exploration, we pursue the question of whether they can create narratives that are indicative of their assigned personality traits. To assess their generated stories, we extract psycholinguistic features using dictionaries that have been designed to analyze human behavior and propose to use these features to study the behavior of \texttt{LLM personas}. We also investigate whether \texttt{LLM personas} are proficient in using lexicons related to their assigned personality profiles and whether they can convincingly portray these personalities to human observers. In other words, can human evaluators discern the assigned personality traits of the \texttt{LLM personas} based on their narratives? Lastly, we suggest potential avenues for extending this evaluation to more real-life scenarios, such as multi-round dialogues and action planning. Motivated by these inquiries, we aim to provide a comprehensive evaluation of \texttt{LLM personas}, focusing on the following research questions (RQs):


\begin{itemize}
    \item \textbf{RQ1}: Can LLMs reflect the behavior of their assigned personality profiles when completing the Big Five Personality Inventory (BFI) assessment?
    \item \textbf{RQ2}: What linguistic patterns are evident in the stories generated by \texttt{LLM personas}?
    \item \textbf{RQ3}: How do humans and LLM raters evaluate the stories generated by \texttt{LLM personas}?
    \item \textbf{RQ4}: Can humans and LLMs accurately perceive the Big Five personality traits from stories generated by \texttt{LLM personas}?
\end{itemize}

\section{Experiment Design}

As shown in Figure \ref{fig:workflow}, this paper investigates the behavior of \texttt{LLM personas} through a multi-faceted approach. We start by creating \texttt{LLM personas} with distinct personality traits and administer a personality assessment to them. Subsequently, we prompt these \texttt{LLM personas} to write stories, which we then analyze using the widely adopted Linguistic Inquiry and Word Count (LIWC) framework. Following this analysis, we recruit human evaluators to manually evaluate the stories and concurrently carry out an LLM-based automatic evaluation. Both human and LLM evaluators are required to (1) assess these stories across six dimensions, namely readability, personalness, redundancy, cohesiveness, likeability, and believability, and (2) infer the personality traits assigned to the LLM personas from the stories. The code, data, and annotations for our experiments are open sourced\footnote{\url{https://github.com/hjian42/PersonaLLM}}.

\subsection{Experiment Setup}

\subsubsection{Model Settings}
We conduct the BFI assessment and LIWC analysis on state-of-the-art LLMs including open-source and closed-source models. We include in the main paper the results for GPT-3.5 (\texttt{GPT-3.5-turbo-0613}) and GPT-4 (\texttt{GPT-4-0613})\footnote{\url{https://platform.openai.com/docs/models}} because our results show that they are more effective at aligning with the designated personas. Results for LLaMA-2 are presented in Appendix \ref{appendix:bfi_llama2} and Appendix \ref{appendix:bfi_scores_liwc}.  Temperature is set as 0.7 to introduce variability in personas' behavior. All other parameters are kept at their default settings.

\subsubsection{\texttt{LLM Persona} Simulation}
For GPT-3.5 and GPT-4, we simulate 10 \texttt{LLM personas} for each combination of the binary Big Five personality types, resulting in 320 distinct personas. They are referred to as \textbf{\texttt{GPT-3.5 personas}} and \textbf{\texttt{GPT-4 personas}} respectively. Figure \ref{fig:workflow} illustrates how we prompt an LLM to generate personas and complete specific tasks. Initially, we create an \texttt{LLM Persona} with a system prompt: ``You are a character who is [TRAIT 1, ..., TRAIT 5].'', where [TRAIT 1, ..., TRAIT 5] represents the assigned Big Five personality. For each personality dimension, we choose one descriptor among the following pairs: (1) extroverted / introverted, (2) agreeable / antagonistic, (3) conscientious / unconscientious, (4) neurotic / emotionally stable, (5) open / closed to experience. 

\subsubsection{BFI Personality Test}
After specifying a personality type, we ask the \texttt{LLM persona} to complete the 44-item Big Five Inventory (BFI), a widely-used self-report scale designed to measure the Big Five personality traits. Only the responses that strictly adhere to the instruction format ``(x) y'' are accepted, where (x) indicates the question number and y indicates the level of agreement on a scale from 1-5. As the green section demonstrates in Figure \ref{fig:workflow}, ``(a) 5'' would indicate that the persona strongly agrees that it talks a lot. Each \texttt{LLM persona}'s responses are aggregated into five personality scores, which are used in later analysis. We use the BFI to assess the personality profiles expressed by LLMs because it is widely utilized in personality-related studies, including many studies involving LIWC. 


\subsubsection{Storywriting}
Subsequently, we prompt these 320 \texttt{LLM personas} to write personal stories with the following simple prompt: ``Please share a personal story in 800 words. Do not explicitly mention your personality traits in the story.'' We impose this restriction to prevent the persona from revealing its hidden attributes, ensuring an unbiased text-based personality assessment by other LLMs and human raters. We tried multiple prompt variants in our initial experiment and decided to purposefully simplify the prompt to reduce demand characteristics for the generalizability of the result. Examples of LLM-generated stories are included in Appendix \ref{appendix:story_examples}.

\subsection{Evaluation Methods}

We evaluate \texttt{LLM personas}' storywriting with a three-pronged analytical approach. First, we conduct a Linguistic Inquiry and Word Count (LIWC) analysis on stories generated by \texttt{GPT-3.5} and \texttt{GPT-4 personas}. Subsequently, we recruit human evaluators and use LLM evaluations to rate these stories from various perspectives. Lastly, we request human evaluators to infer the personality traits of the story author. In human evaluation, the evaluators are randomly assigned to one of two conditions: they are either made aware or kept unaware that the stories were written by an LLM. This study design is to investigate how awareness of AI authorship impacts the evaluation of the narratives and the accuracy of their personality predictions.

Despite explicit instructions to not include any mention of the personality traits, LLMs sometimes failed to follow this instruction. Accordingly, for human evaluation, we sample from stories that do not explicitly mention personality traits to avoid compromised performance in personality prediction. Details of the sampling step are included in Appendix \ref{appendix:lexicon_violation}. With a lexicon-based classifier, we find that most stories produced by \texttt{GPT-3.5 personas} contained explicit references to personality traits (96.56\% compared to GPT-4's 31.87\%). Therefore, we focus on the stories generated by \texttt{GPT-4 personas} in the final human evaluation.


\subsubsection{LIWC Analysis}
We use LIWC-22\footnote{\url{https://www.liwc.app/}} to extract psycholinguistic features from stories generated by \texttt{LLM personas}. By examining the correlation between these features and the personas' assigned personality traits, we aim to identify patterns of linguistic characteristics corresponding to certain personality traits. To compare with human language use, we perform the same analysis on human-generated writing samples from the Essays dataset \cite{pennebaker1999linguistic}\footnote{The Essays dataset, collected from 2,467 participants between 1997 and 2004, consists of stream-of-consciousness essays. Participants also provided self-assessments of the Big Five personality traits in binary form. Note that our personal story prompt differs from the stream-of-consciousness prompt in Essays. However, our comparison aims to approximate the linguistic behavior differences between LLM personas and human writers.} consisting of short essays written by human participants and their self-reported Big-Five personality traits. We then examine whether the linguistic markers associated with certain personality traits are consistent between human and LLM writers.

\subsubsection{Story Evaluation}

We recruit both human and LLM raters to evaluate a subset of the stories generated by \texttt{GPT-4 personas}. Due to budget constraints, we sample 1 out of 10 stories from each personality type, which do not explicitly mention any personality trait (Appendix \ref{appendix:lexicon_violation}). This results in 32 LLM-generated stories for evaluation. For the human evaluation, we recruit five raters to judge each story across six dimensions on a scale of 1 to 5: (1) \textbf{Readability}: whether the story is easy to read, well-structured, and flows naturally, (2) \textbf{Personalness}: whether the story is personal, revealing the writer's thoughts, feelings, and personal experiences, (3) \textbf{Redundancy}: whether the story is concise and free from unneeded content, (4) \textbf{Cohesiveness}: whether sentences in the story fit together well and are logically organized and coherent, (5) \textbf{Likeability}: whether the story is enjoyable or entertaining to read, (6) \textbf{Believability}: whether the story is convincing and realistic, grounded in real-life situations. For the LLM evaluation, we follow \cite{chiang-lee-2023-large} to use GPT-3.5 and GPT-4 evaluators (temperature $=0$) with identical criteria as human raters. The exact prompts given to human and LLM raters are in Appendix \ref{appendix:prolific_setup}.



\subsection{Personality Prediction}
On the same collection of 32 stories, each human annotator and LLM evaluator is asked to predict Big Five personality traits of the writer from the story on a scale of 1 to 5. The objective is to evaluate whether the writing samples from \texttt{LLM personas} can effectively exhibit personality traits to the extent that they are discernible by both human and LLM raters. For each of the personality trait, we provide the descriptions from the work by \citet{john1999big} to the human evaluators as references (see Appendix \ref{appendix:prolific_setup}).

\section{Results}
\subsection{RQ1: Behavior in BFI Assessment}

\begin{figure}[ht!]
    \centering
   \begin{minipage}{0.50\linewidth}
    \centering
    \includegraphics[width=1.0\linewidth]{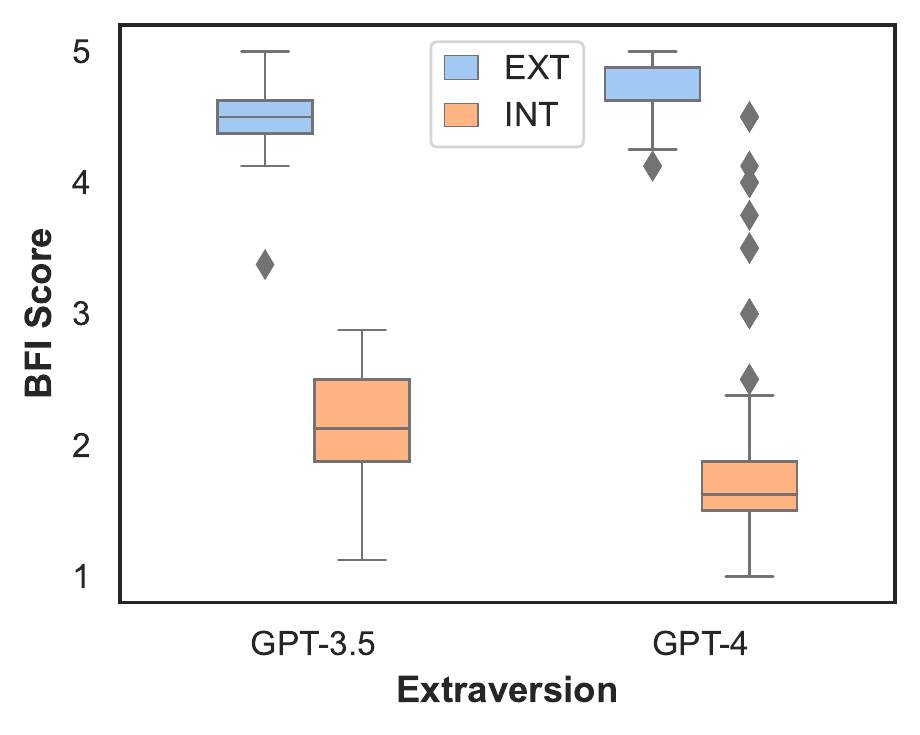}
   \end{minipage}
   \begin{minipage}{0.50\linewidth}
     \centering
     \includegraphics[width=1.0\linewidth]{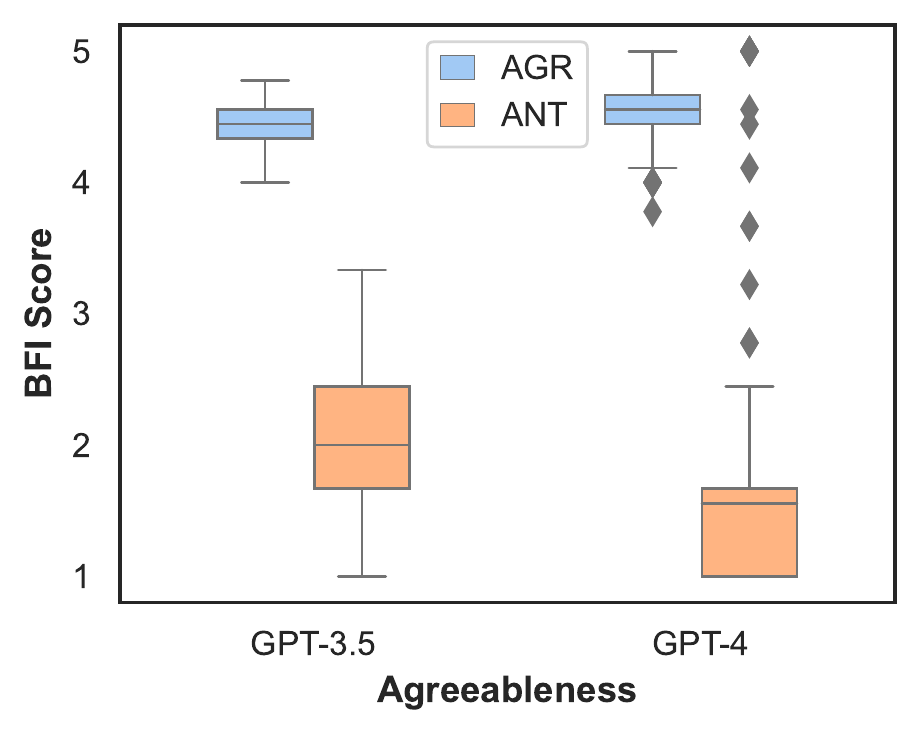}
   \end{minipage}\hfill
   \begin{minipage}{0.50\linewidth}
     \centering
     \includegraphics[width=1.0\linewidth]{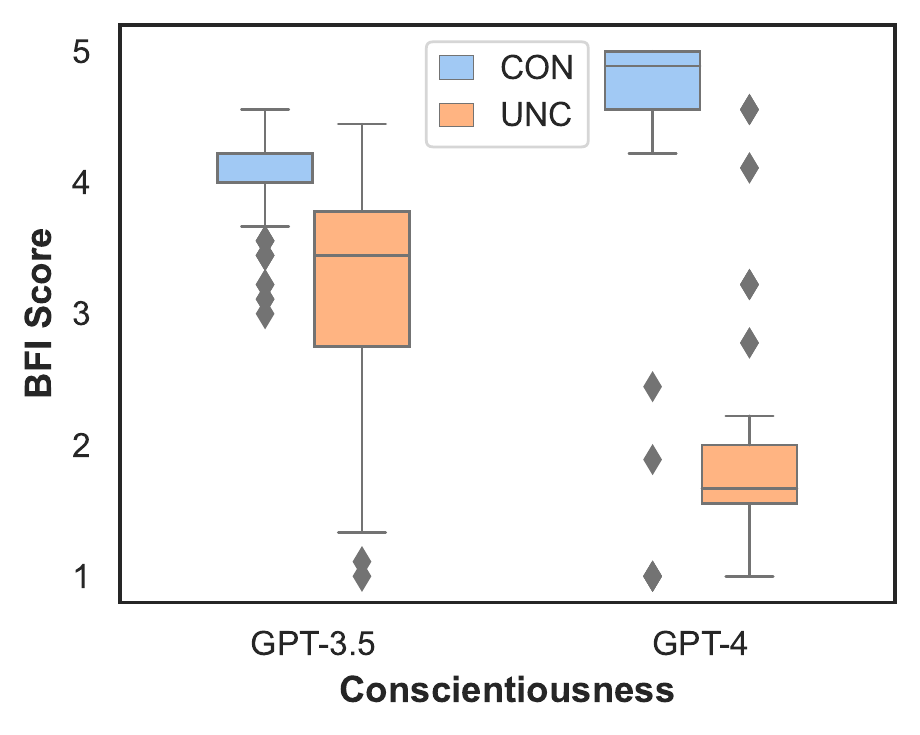}
   \end{minipage}
    \begin{minipage}{0.50\linewidth}
     \centering
     \includegraphics[width=1.0\linewidth]{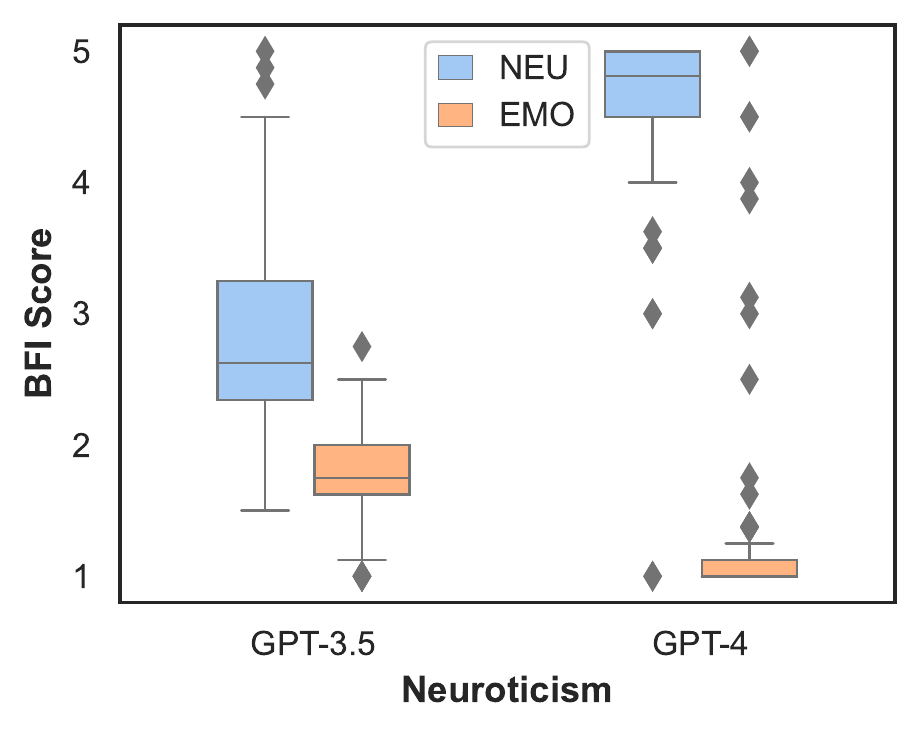}
   \end{minipage}\hfill
   \begin{minipage}{0.50\linewidth}
     \centering
     \includegraphics[width=1.0\linewidth]{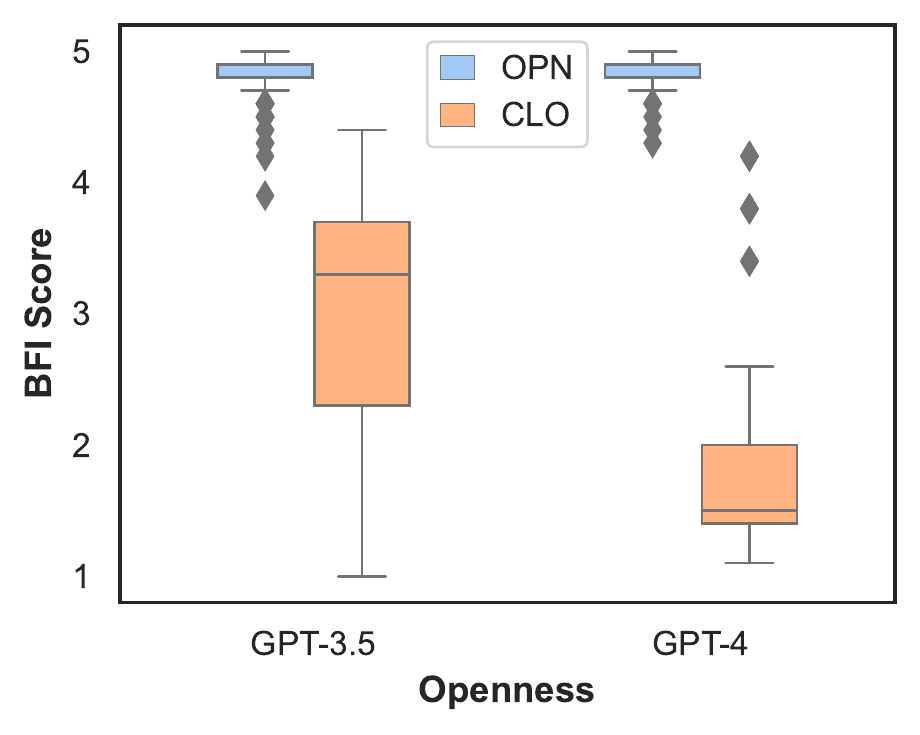}
   \end{minipage}
    \caption{BFI assessment in five personality dimensions by \texttt{GPT-3.5} and \texttt{GPT-4 personas}. Significant statistical differences are found across all dimensions.}
    \label{fig:bfi_stats}
\end{figure}

Based on their responses to the BFI scale, we calculate the personality scores for the 320 GPT-3.5 personas and the 320 GPT-4 personas. We then analyze the distribution of these scores as a function of the assigned personality traits. Specifically, paired t-tests are applied to evaluate the differences between the means of the personality score. The results reveal \textbf{statistically significant differences across all five personality traits}. Large effect sizes are observed for for both \texttt{GPT-3.5 personas} (\textit{EXT}: $d=7.81, p < .001$; \textit{AGR}:d= $5.93, p < .001$; \textit{CON}: $d=1.56, p < .001$; \textit{NEU}: $d=1.83, p < .001$; \textit{OPN}: $d=2.90, p < .001$) and \texttt{GPT-4 personas} (\textit{EXT}: $d=5.47, p < .001$; \textit{AGR}: $d=4.22, p < .001$; \textit{CON}: $d=4.39, p < .001$; \textit{NEU}: $d=5.17, p < .001$; \textit{OPN}: $d=6.30, p < .001$). As shown in Figure \ref{fig:bfi_stats}, BFI scores are lower for LLM personas when they are assigned negative traits (representing the lower end of the trait spectrum) compared to positive traits across all personality dimensions. In summary, the findings effectively address \textbf{RQ1}, substantiating that \textbf{\texttt{LLM personas} are indeed reflecting their assigned personas in BFI assessment}.

\begin{table*}[ht]
\small
\centering
\begin{tabular}{lccccccl}
                     \textbf{Trait} & \textbf{Selected LIWC Features}         & \textbf{Lexicons} & \textbf{GPT-3.5} & \textbf{GPT-4} & \textbf{Humans} & \textbf{GPT-3.5\#} & \textbf{GPT-4\#} \\ \Xhline{3\arrayrulewidth}
\multirow{5}{*}{EXT} & Positive Tone & good, well, new, love & + & + & + & \multirow{5}{*}{\textbf{16/18}} & \multirow{5}{*}{10/18}
\\
                     & Affiliation & we, our, us, help & + & + & + &  \\
                     & Certitude & really, actually, real & - & - &  &  \\
                     & Social Behavior & said, love, care & + & + &  &  \\   
                     & Friends & friend & + &  &  + &  \\  
                     \hline
\multirow{5}{*}{AGR} & Moralization & wrong, honor, judge & - & - & - & \multirow{5}{*}{\textbf{16/23}} & \multirow{5}{*}{13/23} \\
                     & Interpersonal Conflict & fight, attack & - & - & - &  \\
                     & Affiliation & we, our, us, help & + & + & + &  \\
                     & Negative Tone & bad, wrong, hate & - & - & - &  \\
                     & Prosocial Behavior & care, help, thank & + & + &  &  \\ \hline
\multirow{6}{*}{CON} & Drives & we, our, work, us &  & + & + & \multirow{6}{*}{1/31} & \multirow{6}{*}{\textbf{11/31}} \\
                     & Achievement & work, better, best & + & + &  &  \\
                     & Lifestyle (Work, Money) & work, price, market &  & + & + &  \\
                     & Moralization & wrong, honor, judge & - &  & - &  \\
                     & Interpersonal Conflict & fight, attack &  &  & - &  \\
                     & Time & when, now, then &  &  & + &  \\\hline
\multirow{6}{*}{NEU} & Anxiety & worry, fear, afraid & + & + & + & \multirow{6}{*}{7/27} & \multirow{6}{*}{\textbf{15/27}} \\
                     & Negative Tone & bad, wrong, hate & + & + & + &  \\
                     & Mental Health & trauma, depressed & + & + & + &  \\
                     & Sadness & sad, disappoint, cry & - & + & + &  \\
                     & Anger & hate, mad, angry &  & + & + &  \\
                     & Perception (Feeling) & feel, hard, cool &  & + & + &  \\\hline
\multirow{5}{*}{OPN} & Curiosity & research, wonder & + & + & + & \multirow{5}{*}{2/36}  & \multirow{5}{*}{\textbf{17/36}} \\
                     & Insight & know, how, think &  & + & + &  \\
                     & Affiliation & we, our, us, help &  & - & - &  \\
                     & Perception (Visual) & see, look, eye &  & + & + &  \\
                     & Future Focus & will, going to &  & - & - &  \\
                     \Xhline{3\arrayrulewidth}
\end{tabular}
\caption{Correlated metrics between LIWC features and binary personality traits using point-biserial correlation. The analysis is done on personal stories generated by GPT-3.5 and GPT-4 and the human Essays corpus \cite{pennebaker1999linguistic}. This analysis focuses on the psychological and extended vocabulary metrics (81 in total). We report the representative personality LIWC features ($+$ means positive correlation, $-$ means negative correlation) and the \# of overlapped significant LIWC features for GPT-3.5 and GPT-4 with human writings.}
\label{tab:liwc_table}
\end{table*}

\subsection{RQ2: Linguistic Patterns in Writing}

We extract psycho-linguistic features from personal stories generated by \texttt{LLM personas} using LIWC and then calculate point biserial correlations between these features and assigned personality types. The correlation measure is suitable for analyzing the relationship between binary (assigned personality type) and continuous variables (LIWC features). Subsequently, we compare these correlations with those found in human data (i.e., the Essays dataset). 

Table \ref{tab:liwc_table} summarizes the LIWC features that have a statistically significant correlation with certain personality traits. We find that \textbf{assigning different personality types considerably influences the linguistic style of \texttt{LLM personas}}. For instance, for both GPT-3.5 and GPT-4, we find that assigning an \texttt{LLM persona} to be open to experience positively correlates with its use of curiosity lexicons. Similarly, \texttt{GPT-3.5/GPT-4 personas} assigned to be neurotic are more likely to use lexicons related to anxiety, negative tones, and mental health. Also, assigned extraversion correlates positively with lexicons related to positive tone and affiliation. In Appendix \ref{appendix:bfi_scores_liwc}, we include a similar analysis between BFI scores (instead of assigned personality types) and LIWC.

Importantly, these correlations mirror patterns observed in human data (the Essays dataset), indicating \textbf{a notable alignment in word usage between the human dataset and \texttt{LLM personas} writings}. We report the number of shared significant correlations between human and LLM data (denoted as GPT-3.5\# and GPT-4\#) in Table \ref{tab:liwc_table}. \textbf{GPT-4 exhibits greater alignment with humans than GPT-3.}5, with more overlapping lexicons across various traits. This difference is particularly pronounced for Conscientiousness and Openness, where \texttt{GPT-3.5 personas} have $1/31$ and $2/36$ overlapping correlations with humans on Conscientiousness and Openness respectively, whereas \texttt{GPT-4 personas} have $11/31$ and $17/36$. 

\begin{table*}[ht!]
    \centering
    \small
    \begin{tabular}{ccccccc}
        {\textbf{Evaluator}}  & 
        {\textbf{Readability}}  &
        {\textbf{Redundancy}} & 
        {\textbf{Cohesiveness}} & 
        {\textbf{Likability}} &
        {\textbf{Believability}} & 
        {\textbf{Personalness}} \\
        \Xhline{3\arrayrulewidth}
        \multicolumn{7}{c}{\textbf{Uninformed Condition} -- \textit{Evaluation Scores (Mean\textsubscript{STD})}} \\ 
        Human & $4.28_{0.85}$ & $3.70_{1.17}$ & $4.23_{0.88}$ & $3.74_{1.00}$ & $3.96_{1.02}$ & $4.32_{0.85}$\\
        GPT-3.5 & $4.75_{0.43}$ & $3.04_{0.40}$ & $4.97_{0.17}$ & $4.22_{0.48}$ & $3.93_{0.25}$ & $3.55_{0.61}$\\
        GPT-4 & $4.94_{0.24}$ & $4.96_{0.22}$ & $5.00_{0.00}$ & $4.84_{0.36}$ & $4.93_{0.25}$ & $5.00_{0.00}$\\
        \Xhline{3\arrayrulewidth}
        \multicolumn{7}{c}{\textbf{Informed Condition} -- \textit{Evaluation Scores (Mean\textsubscript{STD})}} \\ 
        Human & $4.38_{0.70}$ & $3.62_{1.16}$ & $4.12_{0.82}$ & $3.80_{0.98}$ & $3.97_{0.80}$ & $3.99_{0.90}$\\
        GPT-3.5 & $4.97_{0.17}$ & $2.99_{0.35}$ & $5.00_{0.00}$ & $4.22_{0.41}$ & $3.97_{0.17}$ & $3.31_{0.77}$\\
        GPT-4 & $5.00_{0.00}$ & $4.92_{0.33}$ & $5.00_{0.00}$ & $4.84_{0.36}$ & $4.91_{0.28}$ & $5.00_{0.00}$\\
        \hline
        
    \end{tabular}
    \caption{LLM and human evaluation results of GPT-4 generated stories \textbf{across six dimensions}. \textbf{Uninformed} and \textbf{informed} conditions indicate whether evaluators are informed that the stories are generated by AI. For each attribute, we report its mean Likert scale and the standard deviation. Temperature is set to 0 for both GPT-3.5 and GPT-4.}
    \label{tab:llm_human_eval_scores}
\end{table*}

We further observe in Table \ref{tab:liwc_table} that the stereotypical characteristics of certain personalities might be reflected in LLM linguistic usage while having a different result from the human dataset.
For example, one of the traits associated with high Conscientiousness is achievement striving. This trait is positively correlated with the \texttt{LLM personas}, but does not hold significant correlation in human writings. Furthermore, the emotion of sadness, linked to Neuroticism, shows a negative correlation in writings produced by \texttt{GPT-3.5 personas}. However, it is positively correlated in both \texttt{GPT-4 persona} and human writings, aligning with the typical characteristics of this personality group. Our hypothesis is that LLMs are prone to exhibit strong characteristics of assigned personas while human participants' personalities have much granularity and individual differences. However, it is important to clarify that the human writings from the Essays dataset serve as a comparative reference to gauge the expressivity of LLMs. They should not be considered as an absolute standard, given that human-authored and LLM-generated narratives are not created under identical prompts.

\subsection{RQ3: Story Evaluation}

Next, we extend our analysis to other aspects of the stories generated by  \texttt{GPT-4 personas}, evaluated by both human and LLM raters. Given the subjective nature of the evaluation, we observe a low inter-annotator agreement (IAA) among three annotators, mirroring the findings of \citet{chiang2023can}. The detailed scores can be found in in Appendix \ref{appendix:annotation_agreement}. Consequently, five human or LLM evaluators are recruited for a collective evaluation. In Table \ref{tab:llm_human_eval_scores}, we have the following interesting observations. These stories generated by \texttt{GPT-4 personas} receive high ratings, close to or higher than 4.0, regarding readability, cohesiveness, and believability from both human and LLM evaluators. This suggests that \textbf{the stories are not only linguistically fluent and structurally cohesive, but also convincingly believable}. Furthermore, human evaluators assign high scores for personalness, indicating that these stories genuinely describe personal experiences. Interestingly, these stories receive lower scores for likeability from human evaluators, suggesting that while the stories may be believable and personal, they might not necessarily be as engaging or enjoyable to read. We also discover some interesting comments on these stories from human evaluators (see Appendix \ref{appendix:comments}).

Unsurprisingly, the GPT-4 rater assigns the highest ratings across all dimensions, indicating that \textbf{the GPT-4 rater has a strong preference towards stories generated by GPT-4}. This confirms previous findings that LLMs prefer LLM-generated content \cite{liu2023g}. Notably, the \textbf{GPT-3.5 evaluator assigns lower ratings in redundancy and personalness than both human and GPT-4 evaluators}. We also try multiple temperatures, finding that such trends are consistent in Appendix \ref{appendix:llm_evaluator_check}.


Furthermore, interesting observations are found when the evaluators are informed about the story source, as shown in Table \ref{tab:llm_human_eval_scores}. First, \textbf{human evaluators' perception of stories remains consistent in readability, redundancy, cohesiveness, likeability, and believability regardless of whether they are aware that the content is generated by an LLM}. Second, there is a significant drop in the perceived personalness of the content when human evaluators are informed that the writer is an LLM, suggesting that \textbf{knowledge of the content's origin may influence their sense of connection to the material}. Third, the GPT-3.5 evaluator assigns notably higher ratings for readability and markedly lower ratings for personalness when aware that the content is AI-generated. Finally, the ratings from the GPT-4 evaluator are consistently high with minimal variation between the informed and uninformed conditions, indicating a strong and consistent bias in favor of GPT-4 content.


\subsection{RQ4: Personality Perception}

To assess whether personality traits are predictable from these stories, we undertake two distinct analyses. First, we treat each persona's personality traits as a binary classification problem and compute the accuracy of both humans and LLMs in inferring personality traits. Second, we extract the persona's personality scores and examine the linear relationship between human judgment and the persona's BFI score. A comprehensive overview of the average ratings from humans and LLMs across the five personality dimensions is in Appendix \ref{appendix:personality_ratings}.

\begin{figure*}[ht]
\centering
\begin{minipage}{.48\linewidth}
  \centering
  \includegraphics[width=\textwidth]{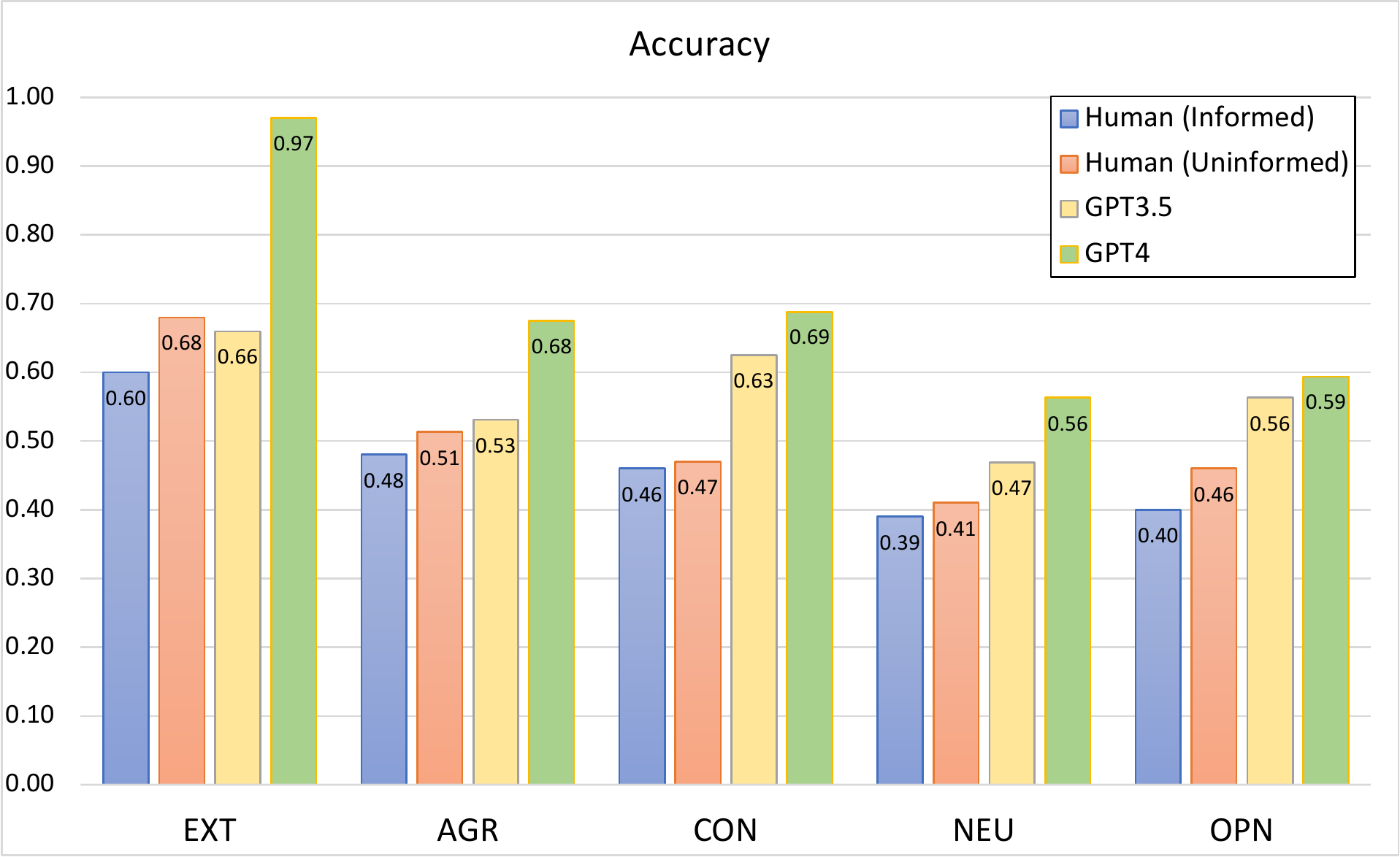}
  \captionof{figure}{\textbf{Individual accuracy} of human and LLM evaluators in predicting personality.}
  \label{fig:accuracy}
\end{minipage}%
\hspace{.5cm}
\begin{minipage}{.48\linewidth}
  \centering
  \includegraphics[width=\textwidth]{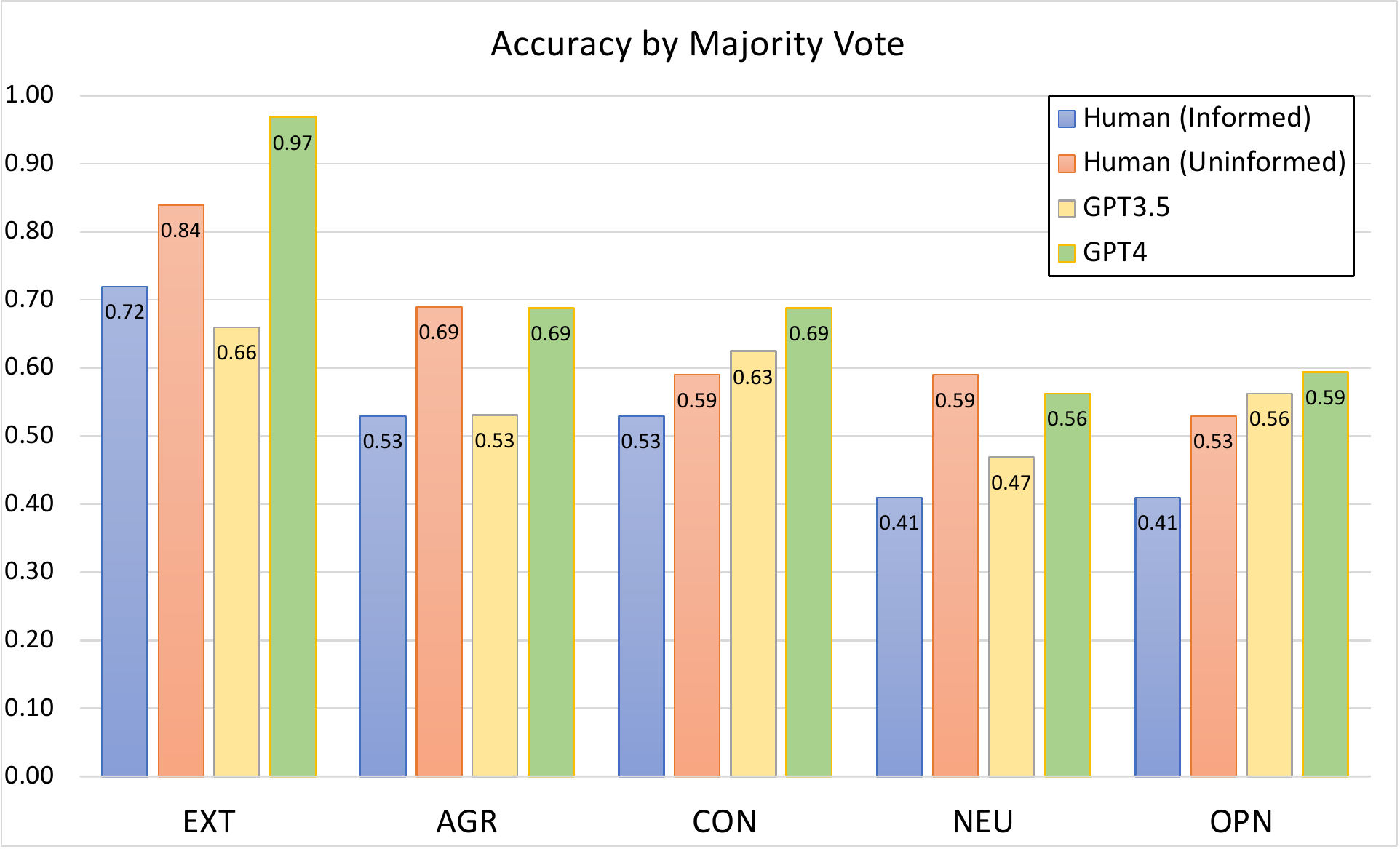}
  \captionof{figure}{\textbf{Collective accuracy} of human and LLM evaluators in predicting personality with majority votes.}
  \label{fig:vote_accuracy}
\end{minipage}
\end{figure*}

\subsubsection{Personality Prediction}

The human evaluators' perceptions of personality were gathered using a Likert scale that ranged from 1 to 5. These numerical values were then transformed into nominal categories. Specifically, scores of 4 and 5 were labeled as ``positive'', 1 and 2 were deemed ``negative'', and a score of 3 was considered ``neutral''. The accuracy of individual and collective evaluations for each story is shown in Figure \ref{fig:accuracy} and Figure \ref{fig:vote_accuracy}, respectively.

The two figures reveal that the accuracy of humans to predict personality traits from stories written by \texttt{GPT-4 personas} varies across the six dimensions. When human evaluators are unaware of AI authorship, they achieve an accuracy of 0.68 on Extraversion and 0.51 on Agreeableness but perform close to random (0.50) on the other BFI dimensions. This shows the difficulty of text-based personality prediction task to human raters. When we aggregate the votes of human annotators based on the majority vote for each story, the accuracy for Extraversion and Agreeableness increases to 0.84 and 0.69, respectively. The accuracy of the other three personality traits also improves with majority voting, indicating \textbf{the personality traits are perceivable (better than random 0.5) from the stories to human raters on a group level}. Interestingly, we find that \textbf{the accuracy decreases with varying degrees when the human evaluators are aware of AI authorship}. Finally, \textbf{GPT-4 shows impressive performance in recognizing Extraversion}, achieving an accuracy of 0.97. GPT-4 also exhibits decent performance in predicting Agreeableness and Conscientiousness, with an accuracy of 0.68 and 0.69, respectively.

\subsubsection{Correlation with BFI Scores}

Furthermore, Spearman's $r$ is calculated between the human's scores for perceived personality trait and the \texttt{Personas} BFI scores on each trait. Our findings reveal that \textbf{\texttt{LLM personas}' BFI scores correlate to varying extents with human perceptions, with Extraversion exhibiting the strongest link}. Specifically, when humans are unaware of AI authorship, significant correlations are found across all five traits (\textit{EXT}: $r=.64, p < .001$; \textit{AGR}: $r=.33, p < .001$; \textit{CON}:  $r=.26, p < .001$; \textit{NEU}: $r=.23, p < .005$; \textit{OPN}: $r=.22, p < .005$). Conversely, when participants knew about AI authorship, correlations persisted in four traits (\textit{EXT}: $r=.42, p < .001$; \textit{AGR}: $r=.32, p < .001$; \textit{CON}:  $r=.20, p < .05$; \textit{NEU}: $r=.17, p < .05$), with non-significance for Openness. The diminished strength of the BFI correlations in the condition where evaluators are informed of AI authorship corroborates our earlier observation: \textbf{the awareness of AI authorship influences the perception of personality}.

\section{Related Work}

\subsection{Personality and Language Use}
Psychologists have developed various personality theories to understand common human traits, including the Big Five \cite{briggs1992assessing,de2000big,goldberg2013alternative}, Sixteen Personality Factors (16PF)  \cite{cattell1957personality,cattell2008sixteen}, and the Myers-Briggs Type Indicator (MBTI) \cite{myers1962myers,myers1985guide}. These theories offer consistent and reliable descriptions of individual differences and have been widely applied in practical contexts such as career planning \cite{schuerger1995career,kennedy2004using,lounsbury2005investigation}, academic achievement \cite{ayers1969study,o2007big,dirienzo2010relationship,kajzer2023exploring}, and relationship compatibility \cite{curran1970analysis,hines2008personality}. Psychometric instruments, such as the BFI \cite{john1999big}, NEO-PI-R \cite{costa2008revised}, and MBTI\textsuperscript{\textregistered}\footnote{\url{https://www.themyersbriggs.com/}}, have been developed based on these theories to represent (quantitatively) personality traits in individuals. Furthermore, research has consistently shown a strong correlation between personality and language use \cite{pennebaker1999linguistic,pennebaker2001patterns,lee2007relations,hirsh2009personality}. \citet{pennebaker2001linguistic} introduced a dictionary LIWC (Linguistic Inquiry and Word Count) to summarize features from human writings and demonstrated their correlation with the Big Five personality traits. While most previous research has focused on language use in humans, our study extends this inquiry to LLMs.

\subsection{LLMs as Simulated Agents} Recent research has shown that, as the size of LLMs increases, LLMs demonstrate emerging abilities as agents \cite{andreas2022language} and exhibit human-like behavior in reasoning \cite{dasgupta2022language,webb2023emergent,binz2023using,aher2023using,wong2023word}, role-playing \cite{wang2023rolellm,shao2023character,wang2023does}, and social science experiments \cite{horton2023large,park2023generative,ziems2023can}. These studies primarily leverage advanced prompting techniques to generate human-like behavior within specific contexts. However, there remains a gap in the literature regarding understanding the abilities of LLM-based agents to exhibit certain personality traits and the effect of said abilities on the linguistic behavior of these agents and human perception towards them.

\subsection{Personality in NLP}
    The NLP community has historically been interested in personality research, including automatic text-based personality prediction \cite{mairesse2007using,feizi2021state,bruno2022personality},  personality prediction from digital footprints \cite{farnadi2013recognising,oberlander2006whose,skowron2016fusing,tadesse2018personality}, and personalized dialogue generation \cite{mairesse2007personage,mairesse2011controlling,zhang2018personalizing,qian2018assigning} including stylistic transfer of personality traits such as formality and politeness \cite{kabbara-cheung-2016-stylistic,jin-etal-2019-imat,madaan-etal-2020-politeness}. With the recent wave of LLM models, research has investigated the use of LLMs for automatic personality prediction \cite{ganesan2023systematic,rao2023can,cao2023chatgpt,yang2023psycot}, assessing the ability of LLMs to express certain personality traits \cite{li2022gpt,pan2023llms,safdari2023personality} and on creating benchmarks for assessing this ability \cite{jiang2022evaluating,wang2023does}, in addition to manipulating personality in LLM content via prompting engineering \cite{karra2022estimating,jiang2022evaluating,jiang2022mpi,caron2022identifying,li2023tailoring}. However, none of the previous work has delved into the linguistic behavior of \texttt{LLM personas} nor the human perception of personality-conditioned LLM content. This study aims to fill that gap by not only examining the linguistic behavior of these personas but also by evaluating their generated content through both human and LLM evaluation. We use story evaluation and personality prediction to offer valuable insights into the capacity of \texttt{LLM personas} to utilize personality-related words and their perception by human evaluators.

\section{Conclusion}
\vspace{-2mm}
In this work, we explore the capability of LLMs (with a focus on GPT-3.5 and GPT-4) to consistently express a personality profile using a well-validated personality scale. Specifically, we investigate the behavior of \texttt{LLM personas} in completing the BFI test and story writing and run analyses with psycholinguistic features, human evaluation, and personality prediction.

Through psycholinguistic analysis, we find that \texttt{LLM personas} from GPT-3.5 and GPT-4 can consistently tailor their BFI answers to match their assigned personalities and write with linguistic features characteristic of those personality traits.  Regarding our investigation on linguistic patterns in writing, we found that each personality trait is associated with  different representative linguistic behavior of LLM personas in writing. We also find a notable alignment in word usage between humans and LLM personas. On LLMs' ability to generate stories conditioned on certain personality profiles, we find that the stories are not only linguistically fluent and structurally cohesive, but also convincingly believable. Moreover, our investigation shows that human evaluators' perception of readability, redundancy, cohesiveness, likeability, and believability remains consistent regardless of whether they are aware that the content is generated by an LLM. We also show that human judges are able to predict personality traits (expressed in the LLM-generated content) with varying degrees across various personality traits. Perhaps, the most interesting finding is that awareness of AI authorship influences the human judges' perception of personality as we notice that the accuracy (of predicting prompted personality traits) decreases (with varying degrees) when human judges are aware of AI authorship.

\section*{Limitations}

\paragraph{Focus on Closed Models} Our study mostly focuses on closed GPT models. We did some preliminary exploration with LLaMA 2 but found its output not suitable for human evaluation. LLaMA 2 repeated highly similar content in generated stories, did not follow instructions closely and explicitly mentioned personality lexicons in generated stories, which undermines our work’s goal. Because LLaMA 2 output was not good and given budget constraints, we decided to pick the best model (GPT-4) for human evaluation.

\paragraph{Data Size} Our dataset is not very large in size, but we believe it provides enough variety and depth for meaningful analysis. Indeed, we generated 10 stories per personality type, resulting in 160 stories per personality trait. While we would have liked to generate a larger number of stories, it would have been costly to recruit human evaluators for such a larger number. Despite the constraints, we made analytical decisions that increase the robustness of our studies. For example, we set the temperature to .7 to introduce more variance to our data. Also, analyses were conducted on the personality-trait level (160 instances per trait) instead of personality-type level (10 instances per type) which provides sufficient sample sizes for the analysis.

\paragraph{Task \& Language Variety} Our work evaluates LLMs in personality assessment and writing settings but does not include more naturalistic settings like human interaction and collaboration of LLM personas. Our study solely focuses on English and does not investigate other languages. 

\paragraph{Evaluation \& Interaction} Since the personality perception task is somehow subjective, future studies should collect data about the human annotator's background with a deeper investigation of the effect of the annotator's personality and background on their personality prediction accuracy. Whether there exists a causal relationship between an annotator's personality and their personality perception towards AI agents could be insightful for artificial agent research. A future step could investigate what fundamental factors contribute to the decrease in personality assignment when humans are aware of AI authorship. It could also be linked with embodied agents to investigate how additional modalities impact the person's perception. A promising future direction would explore how personality influences the action planning of LLM personas. 

\section*{Ethical Considerations}

This study strictly adheres to the ACL Code of Ethics for human experiments and has been granted Exempt status by the Institutional Review Board (IRB). We have conducted our research on the Prolific platform, ensuring compliance with Massachusetts laws by compensating our online annotators at a rate of \$15 per hour. In the interest of transparency and reproducibility, we have included the exact instructions and prompts used in this study in either the paper appendix or the GitHub repository. In the human evaluation, we make sure the stories selected do not contain harmful or offensive text. The evaluators are made aware that their responses will be used exclusively for the study, and no personal identifiers will be collected. We follow the license or terms for use for any research artifact we use in the paper. We follow the terms of use release from OpenAI \footnote{\url{https://openai.com/policies/terms-of-use}} and Licenses from LIWC \footnote{\url{https://www.liwc.app/}}. We use the Essays dataset \cite{pennebaker1999linguistic} solely for our research use. We have checked the content of the generated stories in our paper and ensure that there is no sensitive content in the data. 

Personalized LLMs have demonstrated remarkable abilities in generating human-like content. As these generative agents become increasingly prevalent, it is crucial to consider their potential misuse for harmful purposes, targeting individuals, communities, or entire societies. Personified agents have the potential to provide more enticing interactions for people in their daily lives. Although we do not take a general stance on AI agent applications, we strongly advocate for all stakeholders to disclose their transparency in AI usage to increase trust among individuals. One of our results suggests the necessity of ethical disclosure of AI usage to human users: human's reported personalness and perception of psychological personality traits is greatly impacted by their awareness of AI usage.

Lastly, it is important to emphasize that the primary objective of this work is a scientific inquiry into LLM's expressivity and human personality perception of written records. The evaluation used story writing as a vehicle because it is effective for the purpose of the study and does not have a strong implication for a specific application. We urge all parties to remain vigilant and proactive in mitigating the risks associated with AI to prevent its misuse.

\section*{Acknowledgements}

We would like to thank Matt Groh and Jiangjie Chen for their helpful discussions. Moreover, we thank the reviewers from the International Conference on Computational Social Science (IC2S2) where a preliminary version of this work appeared as a (non-archival) extended abstract. We also thank the reviewers from ACL Rolling Review (ARR) for their constructive feedback.


\bibliography{anthology,custom}

\appendix
\onecolumn

\clearpage

\section{Story Examples Generated by LLM}
\label{appendix:story_examples}
Here, we select a few stories as examples for the audience based on the human-evaluation results in Table \ref{tab:selected_stories}. Specifically, we sample one story from each personality trait that achieved relatively high accuracy in human personality inference results. Notably, due to the unbalanced prediction results on each trait, some selected stories have lower accuracy among $5$ annotators.

{
\small
\begin{longtable}{c|p{0.8\textwidth}}
\hline
Personality              & Story                                                                  \\ \hline
Extroverted (5/5)        & One sunny afternoon, I found myself at the local park, my eyes scanning the horizon, watchful of any interesting activity that might crop up. As an individual who thrives in the company of others, I naturally gravitate towards gatherings and events. So, when I spotted a group of people setting up for a concert, I couldn't help but be intrigued.  I approached the group, a mixture of musicians and technical crew, offering a hand. Their initial surprise quickly transformed into appreciation as my enthusiasm and energy seemed to infect them. I've always been someone who can't resist lending a helping hand, especially if it involves an exciting experience.  As the hours wore on, I found myself assisting with multiple tasks, each more exciting than the last. From helping set up the stage to engaging in a friendly banter with the band members, I was in my element. Although the setting sun indicated that I should be tired, my energy seemed boundless.  However, amidst the excitement, I couldn't shake off a nagging feeling. The concert was to start soon, and the lead guitarist seemed to be nowhere in sight. My worry grew when I overheard the band members...                                 \\ \hline
Introverted (5/5)        & In the quiet solitude of my apartment, I often lose myself in the twisting corridors of my thoughts. The world outside buzzes with life, yet I find comfort in the soft hum of the refrigerator and the rhythmic ticking of the clock. My sanctum is a world of books, music, and art, each holding a mirror to the depth of human experiences.   One particular rainy afternoon, I found myself engrossed in a novel that explored the paradox of human existence. It was then that I received a call from my neighbor, Mrs. Jenkins. Her son had broken his arm and she needed someone to watch over her six-year-old daughter, Lily.   I hesitated, aware that I was about to step outside my comfort zone. But the desperation in Mrs. Jenkins' voice nudged me into agreeing. Lily was a whirlwind of energy, her infectious laughter and inquisitive nature a stark contrast to my usual tranquility.   As soon as Lily stepped into my apartment, her eyes widened at the sight of my overflowing bookshelves, my sketches littered on the coffee table, and the classical music playing softly in the background. I could see her mind buzzing with...                                                                                \\ \hline
Agreeable (5/5)          & Once upon a time, I was working as an event coordinator for a large corporation. While I was always busy, I relished the opportunity to be surrounded by a bustling environment and a sea of people. My days were filled with planning meetings, coordinating logistics, and ensuring that everything ran like clockwork.   One day, I was tasked with planning our most significant event of the year, the annual gala. The event was a considerable undertaking, with hundreds of guests, including our CEO, high-profile clients, and notable industry personalities. The pressure was on, but I was ready to rise to the occasion.  I meticulously planned every single detail, from the venue layout to the minute-by-minute schedule. I triple-checked everything, ensuring that the caterers understood the menu, the decorators knew the theme, and the audiovisual team was prepared. My team and I worked tirelessly around the clock to make sure that the event would be perfect.  A week before the gala, I started feeling a knot of unease in my stomach. I was constantly replaying every detail in my head, worrying about what could go wrong. I couldn't shake off the feeling that I was forgetting something crucial.... \\ \hline
Antagonistic (4/5)       & There I was, in the back corner of the local coffee shop, my sanctuary from the world. My gaze was transfixed on the dark, bitter coffee in my hand, the steam rising in wisps like my thoughts. The world outside seemed to buzz and hum, but I was insulated from it all, safe in my solitude.  The bell on the front door jingled, signaling a new customer's arrival. I didn't bother to look up. Why would I? I was there for the coffee, not the people. But, the universe seemed to have other plans.  A voice, chipper and bright, echoed through the shop, cutting through the usual din of murmuring voices and clinking cups. "Hey, can I sit here? All the other seats are taken."  I looked up to see a girl, probably in her early twenties, standing at my table. She had a wide smile that seemed to shine brighter than the sun outside. I gave her a curt nod, gesturing to the empty seat across from me. She sat down, her eyes sparkling with an eagerness that I found unsettling.  She launched into a conversation, talking about the weather, the coffee, the book she...                                                                                                                                            \\ \hline
Conscientious (5/5)      & It was a typical Tuesday afternoon and I was comfortably settled in my favorite corner of the local library. Surrounded by the faint smell of old books and the hushed whispers of other patrons, I felt a sense of calm wash over me. The library was my sanctuary, a place where I could retreat from the world and lose myself in the quiet contemplation of the written word.  One particular book had caught my interest that day, 'The Art of Solitude.' It was a thought-provoking exploration of the joys and challenges of being alone, a theme that resonated deeply with me. As I turned each page, I found myself nodding in agreement with the author's reflections on the importance of solitude in fostering self-awareness and introspection.  A sudden burst of laughter interrupted my reading. Looking up, I saw a group of teenagers huddled around a table, their faces lit up by the glow of their smartphones. I felt a pang of annoyance. This was a library, after all, a place of quiet and respect. But I quickly reminded myself that everyone has a different way of engaging with the world. I took a deep breath, refocused, and immersed...                                                                   \\ \hline
Unconscientious (2/5)    & I remember the time when I was given the responsibility of organizing our company's annual retreat. I was thrilled at the opportunity, not because I enjoy coordinating such events, but because I knew it was going to be a great chance to stir things up a bit.   We were a group of 20 people who would be spending an entire weekend at a remote resort. The retreat was meant to foster team bonding and camaraderie, but honestly, these events were always predictable and monotonous. I decided this one would be different.  I strode into the planning meeting with an air of confidence that was hard to ignore. Most of my colleagues were looking forward to a well-organized and systematic event. They knew I wasn't exactly diligent when it came to planning, but they probably hoped I would make an exception this time.  I had a different plan. I decided to make the retreat as spontaneous and unpredictable as possible. No schedules, no pre-planned activities, just a free-for-all. I stated my plan to the team, and as expected, it was met with raised eyebrows and skeptical looks. I shrugged off their doubt. After all, I wasn't one...                                                                    \\ \hline
Neurotic (3/5)           & My life has always been a tapestry of solace and solitude, a universe of its own. On a chilly winter night, I sat alone in my dimly lit room, the faint glow from my computer screen illuminating the scattered books and papers. I had a deadline to meet, a story to submit, but my mind was adrift, entangled in a web of thoughts, each more intricate than the last. A cup of cold coffee lay discarded on the desk, its purpose long forgotten.  My gaze fell on the worn, leather-bound book peeking from the pile, "Art: The Definitive Visual Guide." A surge of curiosity urged me to pick it up. I consider myself an eclectic reader, my interests spanning from the realms of science to the oasis of arts, and everything in between. I found solace in the unknown, in the novelty of new ideas, and the thrill of uncharted territories.  I opened the book, and the artistic world unfolded before me. The vivid colors, the passionate strokes, and the silent stories captivated me. Each painting was a universe of its own, a tale of the artist's soul, and a mirror reflecting the observer's psyche. The world around...                                                                                              \\ \hline
Emotionally Stable (5/5) & In the heart of the bustling city, I lived in a small apartment, my sanctuary from the world. One day, as I was returning from my weekly trip to the local bookstore, I found a stray kitten huddled by the staircase of my building. It was a tiny, shivering furball, looking up at me with wide eyes. I couldn't ignore it.   I brought the kitten home, creating a warm makeshift bed from an old sweater. I named her Luna. Luna was timid and wary of her new surroundings. But with time, she grew comfortable around me, her purrs filling my once quiet apartment with life.   Being an editor for a publishing house, I worked from home, which meant I spent a great deal of time with Luna. I would read her passages from the manuscripts I was working on, and she would purr in response, as if she understood.   One day, I received a manuscript from a new author. It was a fantasy novel, filled with magical creatures, enchanted forests, and heroic quests. As I read the manuscript, I noticed it was riddled with plot holes and inconsistencies. But beneath the imperfections, there was...                                                                                                                         \\ \hline
Open (5/5)               & There I was, standing at the entrance of the art museum, my heart pounding with anticipation. I was about to experience the grand opening of the museum's new modern art exhibit, a highly anticipated event in our city. I was lucky to have scored an invite to the opening night, and I was buzzing with excitement. I had no idea that this evening would lead me to a new friend and an unforgettable experience.  As I walked into the grand hall, I was greeted by a sea of vibrant colors, unusual shapes, and thought-provoking installations. Each piece of art seemed to tell a unique story, inspiring a sense of curiosity within me. I was eager to explore more, to dive into the unknown, and to allow the art to speak to me.  As I was studying a beautiful abstract painting, a lady standing next to me started a conversation. "Quite a mesmerizing piece, isn't it?" she asked, her eyes sparkling with enthusiasm. She introduced herself as Grace, a local art enthusiast. We instantly clicked and found ourselves engrossed in deep conversations about the various art pieces. We both loved the same paintings, shared the same perspectives, and could...                                                        \\ \hline
Closed (4/5)             & I remember it was a chilly Saturday in December. The sky was overcast, a canvas of grey, and the streets were bustling with last-minute holiday shoppers. I, however, was tucked away in my quiet corner of the world, my small yet cozy apartment.  My place could be described as a bit messy. Books were strewn haphazardly on my coffee table, and my laundry hamper was positively overflowing. But I liked it that way‚Äîit felt lived in, familiar.  That day, I had chosen to retreat from the world and indulge in my favorite pastime: reading. As I sank into my worn-out couch, a worn-out paperback in my hands, I felt a wave of contentment wash over me. There was something comforting about the world of literature. The characters never asked too much of me, and I was free to explore their lives without the pressure of social interaction.  There was a knock at the door, breaking the peaceful silence of my sanctuary. It was my neighbor, Mrs. Jenkins, her arms full of freshly baked cookies. She was an extrovert, always popping by to chat or share her latest culinary creations. Despite our stark personality differences, we had formed...                                                              \\ 
\hline
\caption{Selected stories from human-evaluation experiment. The personality columns shows its corresponding predicted personality accuracy rate. $x/5$ means $x$ out of $5$ annotators predicted correctly.}
\label{tab:selected_stories}
\end{longtable}
}

\section{Story Comments}
\label{appendix:comments}

During human annotation, we provide an optional comment section for each story, allowing human annotators to share their thoughts after reading the story. We receive some interesting comments from human annotators when they are informed or uninformed that the writer of the story is an LLM. After filtering out comments such as ``N/A'', ``No'', ``None'' and ``No Comments'', there are 104 valid comments whose average length is 12.7 words in the informed condition and 122 valid comments whose average length is 13.4 words in the uninformed condition. We compute the sentiment scores of the comments with \texttt{cardiffnlp/twitter-roberta-base-sentiment-latest} by mapping negative, neutral, and positive to -1, 0, and 1. We find out that the average sentiment is $0.45$ in the informed condition and $0.16$ in the uninformed condition, indicating that evaluators have a slightly better attitude towards the stories when informed of AI authorship. We also sample a few representative comments for each condition in Table \ref{tab:comments}. We notice that there are constant mixed comments towards these stories, where some stories are quite believable and enjoyable and other stories are banal and exaggerating. However, we observed that when informed of AI authorship people tend to be more lenient about the stories and give more complements about the stories (see comments in the ``surprised'' section), which is consistent with the higher average sentiment score we show earlier. When people are unaware that the author is an LLM, they constantly guess the author's personality and question the motivation of the author to write some stories (see comments in the ``confused'' section). This is particularly interesting and highlights the potential social implications in terms of confusion if AI-generated content is consumed or AI characters interact with humans without notifying people.  All the comments will be published along with the stories and code on Github. 

\begin{longtable}[ht!]{lp{0.8\textwidth}}
        {\textbf{Attitude}}  & 
        {\textbf{Comments}} \\

        \Xhline{3\arrayrulewidth}
        \multicolumn{2}{c}{\textit{\textbf{Informed Condition}}} \\ 
        \textbf{Critical} & (1) I thought the story was a little basic and lacked deeper meaning. \\
        & (2) The build up is pretty good but falls flat towards the end. \\
        & (3) A believable story, if it seems somewhat exaggerated with the author’s impulsiveness. \\ \hline
        \textbf{Sympathetic} & (1) It's a relatable situation that someone could get behind and feel for. \\
        & (2) Relatable work \& anxiety story. I relate in some ways. \\
        & (3) live your dreams even when there no planned. \\ \hline
        \textbf{Positive} & (1) I found myself really being put into the characters shoes. \\ 
        & (2) I found this story to be hilarious.  I was laughing while I was reading it.  The main character's interactions were absolute greatness. \\
        & (3) Vivid descriptions, almost cinematic like a movie script. \\ \hline
        \textbf{Surprised} & (1) The story actually sounded genuine and I wouldn't have believed it was written by AI unless someone told me.\\
        & (2) I would have never guessed this was an AI writer. I'm quite impressed and I thoroughly enjoyed this road trip story. \\
        & (3) I like this story here the cat was a separate side story but the AI was able to integrate it in throughout, in other stories it had what I would call side quests that added nothing to the final flow. \\
        \hline

        \Xhline{3\arrayrulewidth}
        \multicolumn{2}{c}{\textit{\textbf{Uninformed Condition}}} \\ 
        \textbf{Critical} & (1) It was harder to read and follow the story. \\
        & (2) Some of the punctuation seemed a little odd or over used. \\
        & (3) It started off strong, but there was no sense of why this person became uneasy, and they seemed to become a different person as the essay went on, all of a sudden wanting predictability and alone time instead of the chaos and "social butterfly" status. I found it inconsistent. \\ \hline
        \textbf{Sympathetic} & (1) Also feels quite personal. \\
        & (2) When you constantly look down, you don't see what is right in front of you. \\
        & (3) As someone who used to code a lot, I felt the ending was very moving and believable. \\ \hline
        \textbf{Positive} & (1) Very enjoyable story about how sometimes unavoidable changes in our lives can lead to happier lives. \\ 
        & (2) It was enjoyable. I appreciated the self-awareness by someone who knows they are not always well-liked or well-received. \\
        & (3) I enjoyed the description of the old lady particularly the etchings on her face. This was quite a memorable explanation. \\ \hline
        \textbf{Confused} & (1) Here's something about the story that doesn't seem believable, but it's probably just the writer's extraggretions. \\
        & (2) This person must have taken drugs lol. \\
        & (3) The writer seems like he is not really fun to be around. \\ \Xhline{3\arrayrulewidth}
    \caption{We sample story comments for both informed and uninformed conditions.}
    \label{tab:comments}
\end{longtable}

\section{Personality Ratings}
\label{appendix:personality_ratings}

We sampled 32 \texttt{LLM personas} from 32 personality types. Therefore, we have 16 personas with positive labels and 16 personas with negative labels for each personality, which would ideally lead to the average ratings close to 3. As shown in Table \ref{tab:llm_human_eval_scores_personality}, we find that the average ratings of GPT-3.5 and GPT-4 are closer to 3 than humans in Extraversion. Except for Extraversion, the average ratings from GPT-4 seems consistently further away from 3 compared to human and GPT-3.5 evaluators.

\begin{table*}[h!]
    \centering
    \small
    \begin{tabular}{cccccc}
        {\textbf{Evaluator}}  & 
        {\textbf{Extraversion}}  &
        {\textbf{Agreeableness}} & 
        {\textbf{Conscientiousness}} & 
        {\textbf{Neuroticism}} &
        {\textbf{Openness to Experience}} \\
        \Xhline{3\arrayrulewidth}
        \multicolumn{6}{c}{\textbf{Uninformed Condition} -- \textit{Evaluation Scores (Mean\textsubscript{STD})}} \\ 
        Human & $3.22_{1.36}$ & $3.42_{1.15}$ & $3.86_{1.05}$ & $2.76_{1.22}$ & $3.54_{1.19}$ \\
        GPT-3.5 & $3.19_{0.85}$ & $4.08_{0.82}$ & $3.39_{0.74}$ & $2.13_{0.49}$ & $3.62_{0.60}$ \\
        GPT-4 & $3.00_{1.42}$ & $4.01_{1.08}$ & $4.04_{1.18}$ & $2.02_{1.01}$ & $4.03_{1.07}$ \\
        \Xhline{3\arrayrulewidth}
        \multicolumn{6}{c}{\textbf{Informed Condition} -- \textit{Evaluation Scores (Mean\textsubscript{STD})}} \\ 
        Human & $3.29_{1.17}$ & $3.67_{0.84}$ & $3.76_{0.92}$ & $2.69_{1.23}$ & $3.70_{1.00}$ \\
        GPT-3.5 & $3.14_{0.86}$ & $4.16_{0.91}$ & $3.56_{0.71}$ & $2.03_{0.47}$ & $3.66_{0.59}$ \\
        GPT-4 & $3.00_{1.42}$ & $4.22_{1.09}$ & $4.22_{1.14}$ & $2.02_{1.02}$ & $4.09_{1.08}$ \\
        \hline
        
    \end{tabular}
    \caption{LLM and human evaluation results of GPT-4 generated personal stories in \textbf{5 personality traits}. \textbf{Uninformed} and \textbf{informed} conditions indicate whether human or LLM evaluators are informed that the stories are generated by an LLM. We report each evaluated attribute's mean Likert scale and standard deviation. Temperature is set to 0 for both GPT-3.5 and GPT-4.}
    \label{tab:llm_human_eval_scores_personality}
\end{table*}

\begin{table*}[ht!]
    \centering
    \small
    \begin{tabular}{ccccccc}
        {\textbf{Evaluator}}  & 
        {\textbf{Readability}}  &
        {\textbf{Redundancy}} & 
        {\textbf{Cohesiveness}} & 
        {\textbf{Likability}} &
        {\textbf{Believability}} & 
        {\textbf{Personalness}} \\

        \Xhline{3\arrayrulewidth}
        \multicolumn{7}{c}{\textit{Inter-Annotator Agreement (IAA\textsubscript{\%})}} \\ 
        Uninformed Human & $0.05_{62}$ & $-0.03_{48}$ & $0.03_{61}$ & $0.02_{54}$ & $-0.03_{51}$ & $-0.02_{60}$\\
        Informed Human & $0.01_{64}$ & $0.02_{53}$ & $0.03_{58}$ & $0.06_{55}$ & $-0.02_{57}$ & $0.10_{61}$ \\ \hline
        
    \end{tabular}
    \caption{We report the inter-annotator agreement (IAA) among five annotators across \textbf{six different metrics} using Krippendorff’s $\alpha$. The subscript in the IAA column (\%) is used to denote the average percentage of annotators who agree on the most voted rating.}
    \label{tab:iaa_human_six_metrics}
\end{table*}

\begin{table*}[ht]
    \centering
    \small
    \begin{tabular}{ccccccc}
        {\textbf{Evaluator}}  & 
        {\textbf{Extraversion}}  &
        {\textbf{Agreeableness}} & 
        {\textbf{Conscientiousness }} & 
        {\textbf{Neuroticism}} &
        {\textbf{Openness to Experience}} \\

        \Xhline{3\arrayrulewidth}
        \multicolumn{6}{c}{\textit{Inter-Annotator Agreement (IAA\textsubscript{\%})}} \\ 
        Uninformed Human & $0.11_{51}$ & $0.03_{49}$ & $0.00_{51}$ & $-0.03_{43}$ & $0.04_{52}$ \\
        Informed Human & $0.10_{54}$ & $0.11_{65}$ & $0.07_{59}$ & $0.03_{49}$ & $0.08_{57}$ \\ \hline
        
    \end{tabular}
    \caption{We report the inter-annotator agreement (IAA) among five annotators across \textbf{five personality traits} using Krippendorff’s $\alpha$. The subscript in the IAA column (\%) is used to denote the average percentage of annotators who agree on the most voted rating.}
    \label{tab:iaa_human_five_trait}
\end{table*}

\section{Story Evaluation Details}
\label{appendix:annotation_agreement}

\subsection{Filtering Stories for Evaluation}
\label{appendix:lexicon_violation}

We design a simple lexicon-based classifier to detect if a story contains explicit use of personality trait lexicons. These lexicons include ``extrover*'', ``introver*'', ``agreeabl*'', ``antagonis*'', ``*conscientious*'', ``neuroti*'', ``emotionally stabl*'', ``open to experience'', ``closed to experience''. We filter out stories which contain these lexicons and sample from the remaining stories for human evaluation.

\subsection{Prolific Setup}
\label{appendix:prolific_setup}
We recruit Prolific workers from the United States, whose first language is English with an approval rate between 99\% and 100\%. We have divided 32 stories into four equal batches, each containing eight stories. To begin each batch, a consent form is provided. Following this, each annotator reads the story and answers six evaluation questions that assess readability, personalness, redundancy, cohesiveness, likeability, and believability. An optional comment section is also provided for additional feedback on the story. Subsequently, we ask the annotators five questions related to personality traits: Extraversion, Agreeableness, Conscientiousness, Neuroticism, and Openness to Experience. Screenshots of these questions are included below.

\begin{figure}[ht!]
    \centering
    \includegraphics[width=0.7\linewidth]{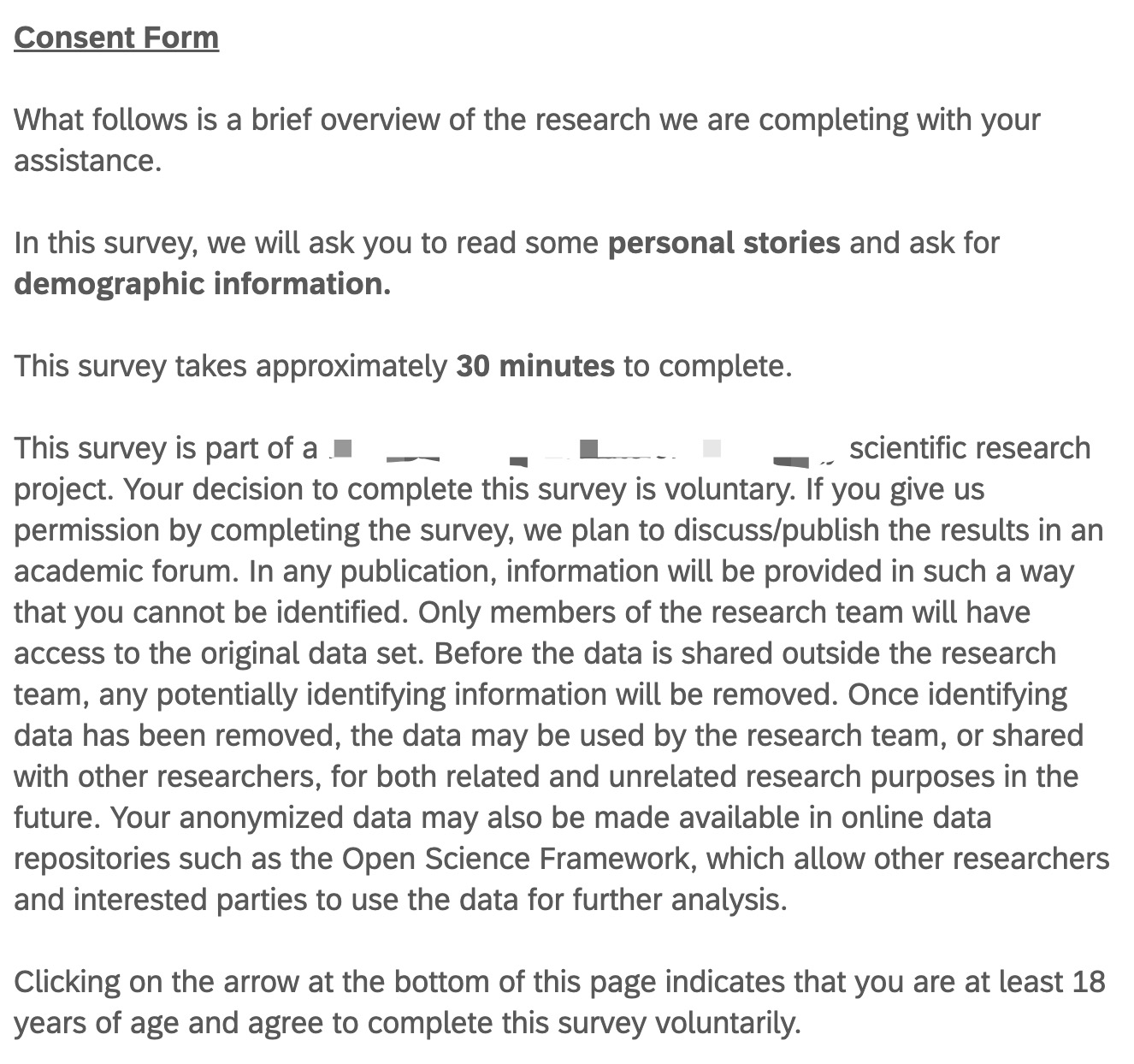}
    \label{fig:consent}
    \caption{Consent form on Prolific.}
\end{figure}

\begin{figure}[ht!]
    \centering
    \includegraphics[width=0.7\linewidth]{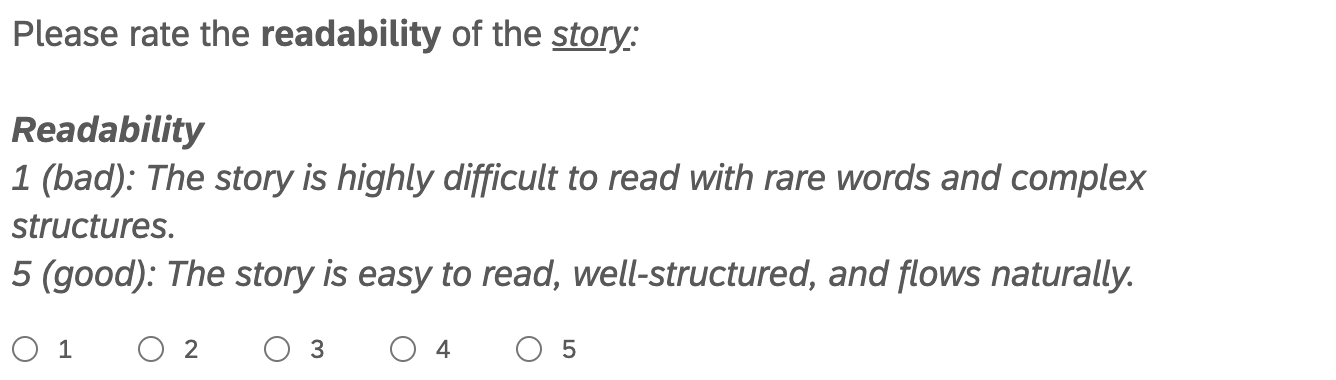}
    \label{fig:readability}
    \caption{Readability question on Prolific.}
\end{figure}

\begin{figure}[ht!]
    \centering
    \includegraphics[width=0.7\linewidth]{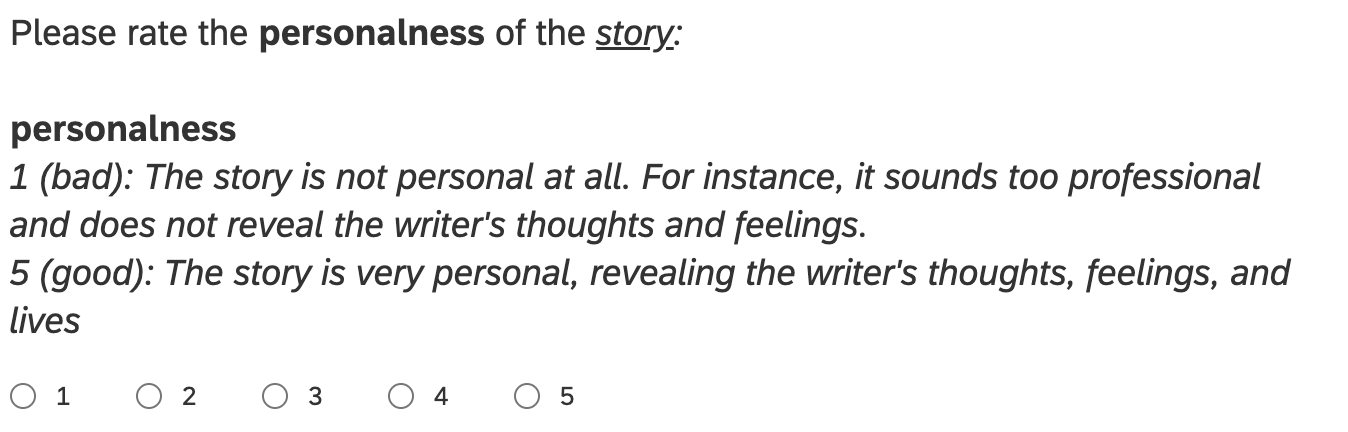}
    \label{fig:personalness}
    \caption{Personalness question on Prolific.}
\end{figure}

\begin{figure}[ht!]
    \centering
    \includegraphics[width=0.7\linewidth]{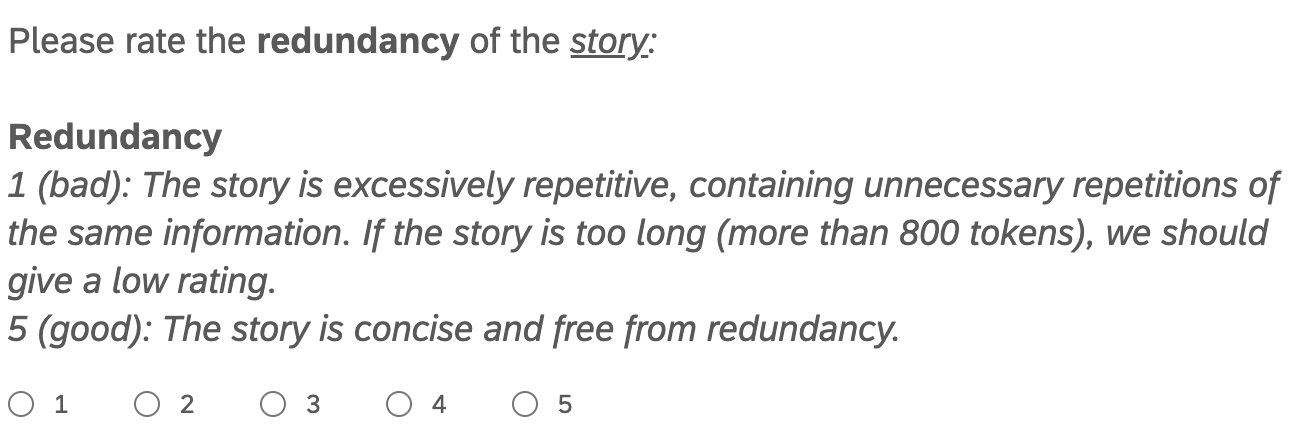}
    \label{fig:redundancy}
    \caption{Redundancy question on Prolific.}
\end{figure}

\begin{figure}[ht!]
    \centering
    \includegraphics[width=0.7\linewidth]{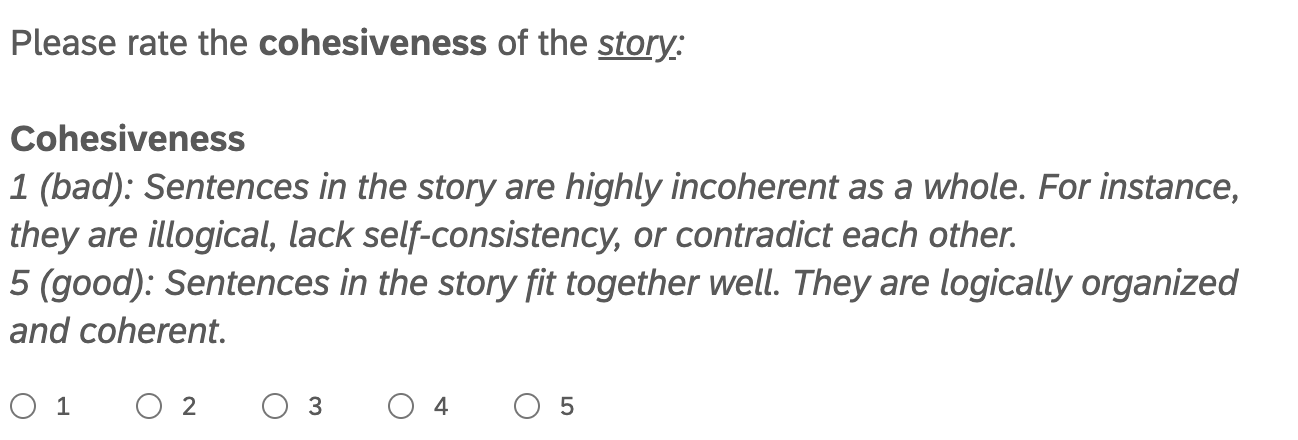}
    \label{fig:cohesiveness}
    \caption{Cohesiveness question on Prolific.}
\end{figure}

\begin{figure}[ht!]
    \centering
    \includegraphics[width=0.7\linewidth]{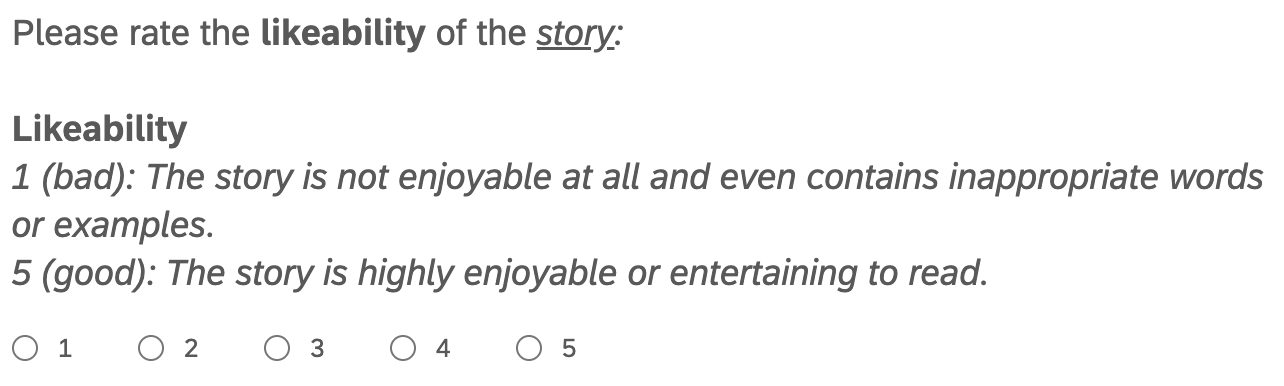}
    \label{fig:likeability}
    \caption{Likeability question on Prolific.}
\end{figure}

\begin{figure}[ht!]
    \centering
    \includegraphics[width=0.7\linewidth]{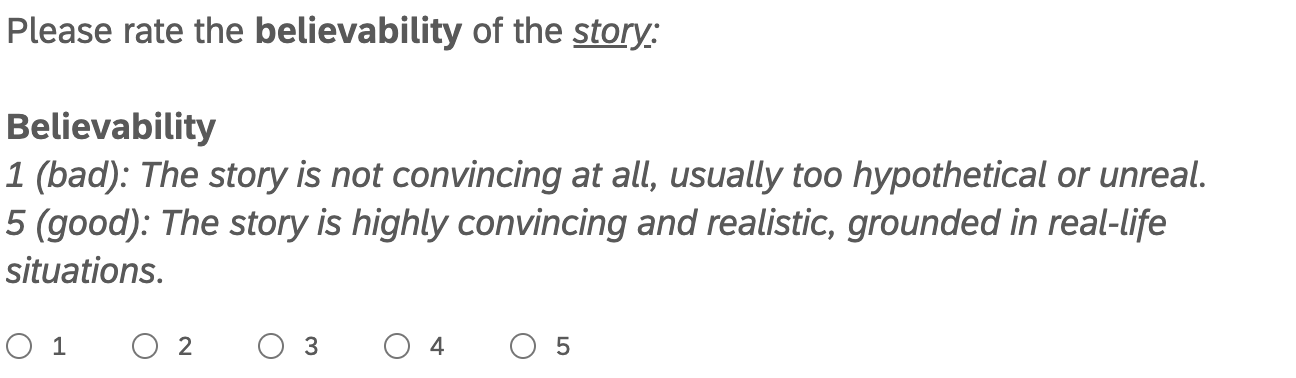}
    \label{fig:believability}
    \caption{Believability question on Prolific.}
\end{figure}

\begin{figure}[ht!]
    \centering
    \includegraphics[width=0.7\linewidth]{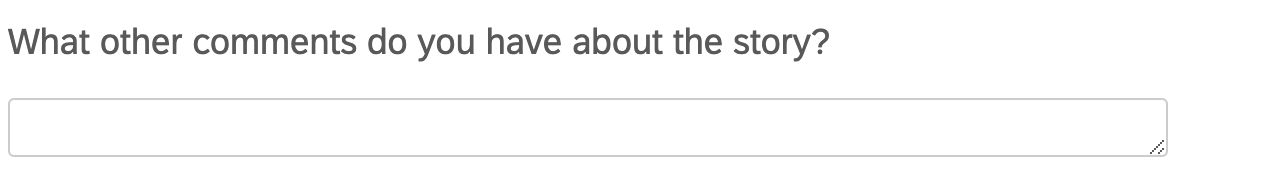}
    \label{fig:comment}
    \caption{Comment question on Prolific.}
\end{figure}

\begin{figure}[ht!]
    \centering
    \includegraphics[width=0.7\linewidth]{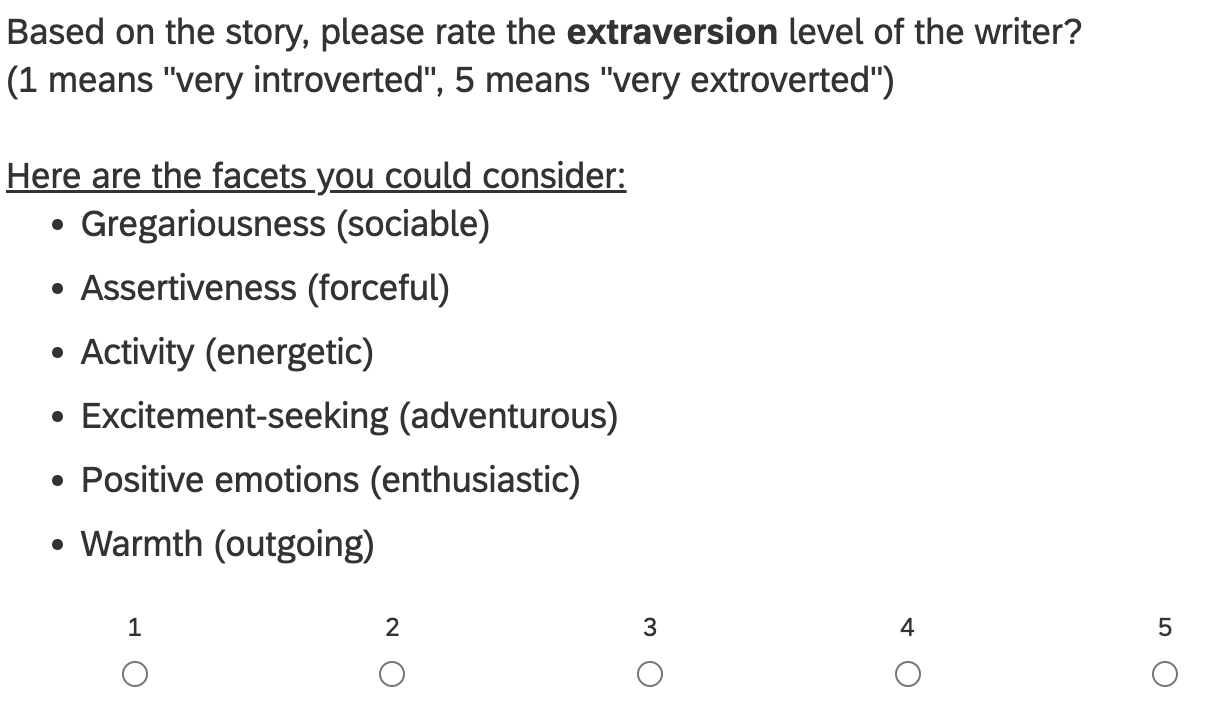}
    \label{fig:extraversion}
    \caption{Extraversion question on Prolific.}
\end{figure}

\begin{figure}[ht!]
    \centering
    \includegraphics[width=0.7\linewidth]{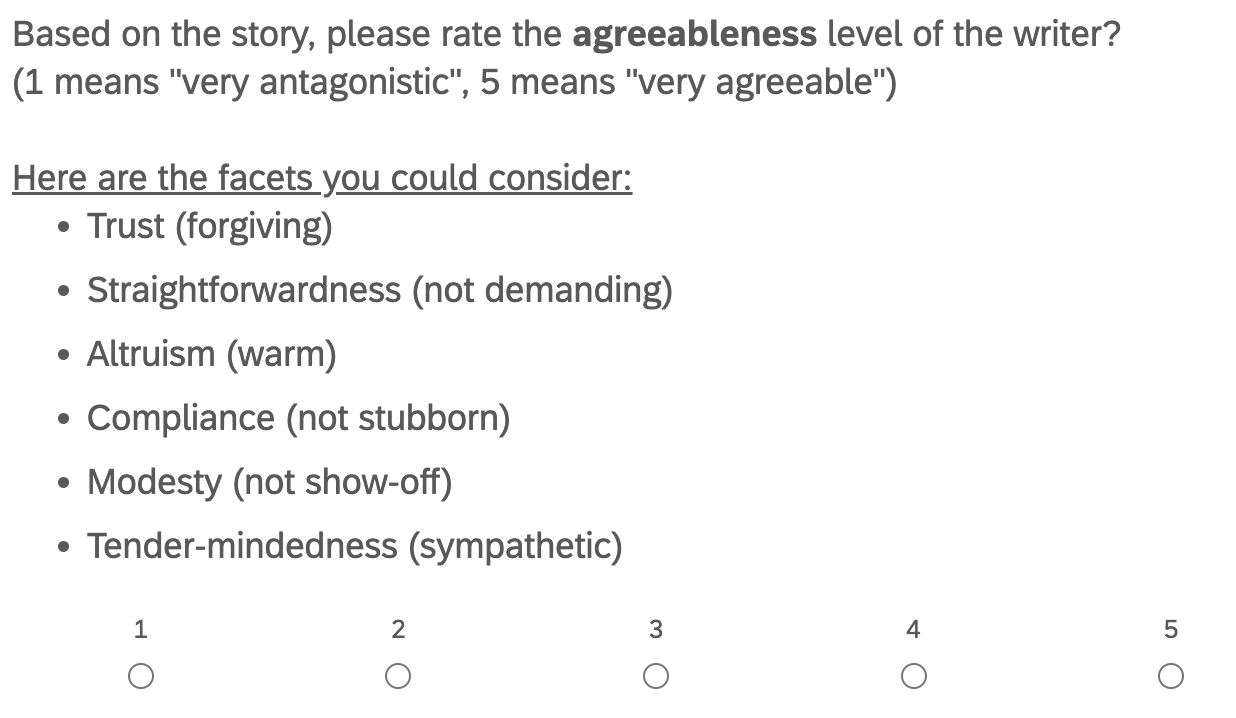}
    \label{fig:agreeableness}
    \caption{Agreeableness question on Prolific.}
\end{figure}

\begin{figure}[ht!]
    \centering
    \includegraphics[width=0.7\linewidth]{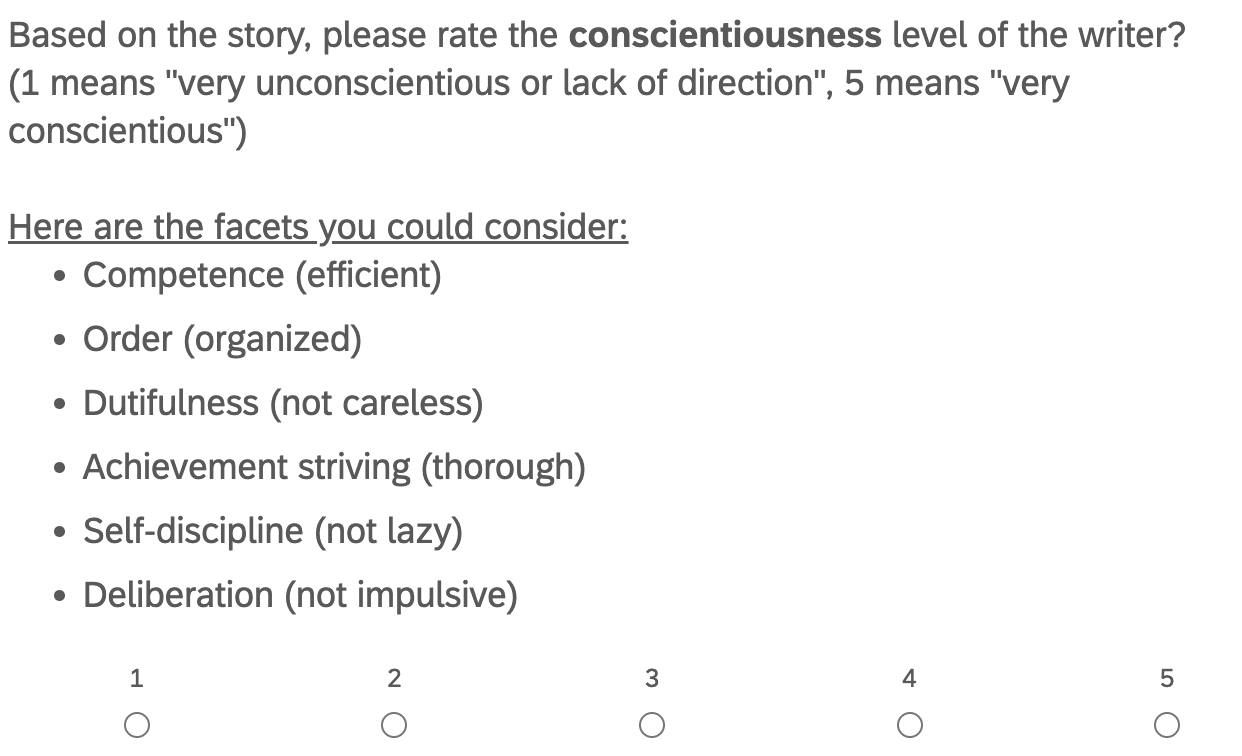}
    \label{fig:conscientiousness}
    \caption{Agreeableness question on Prolific.}
\end{figure}

\begin{figure}[ht!]
    \centering
    \includegraphics[width=0.7\linewidth]{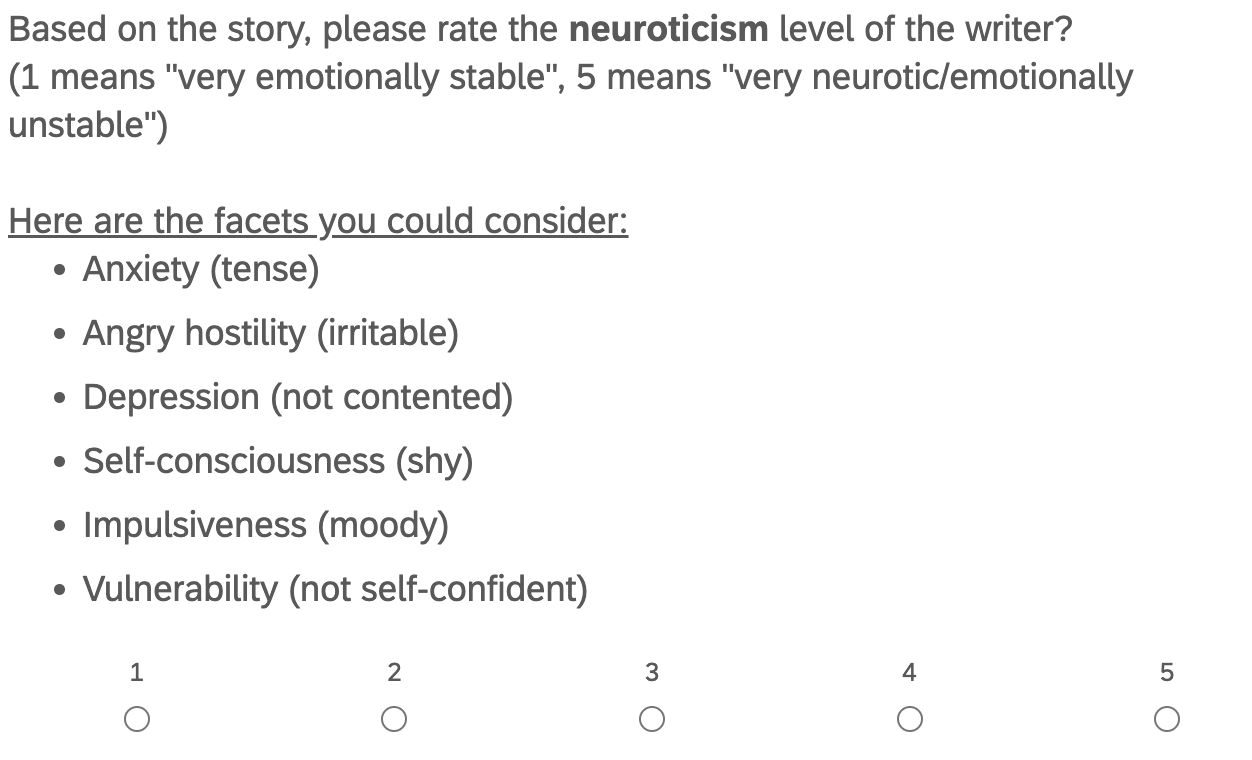}
    \label{fig:neuroticism}
    \caption{Neuroticism question on Prolific.}
\end{figure}

\begin{figure}[ht!]
    \centering
    \includegraphics[width=0.7\linewidth]{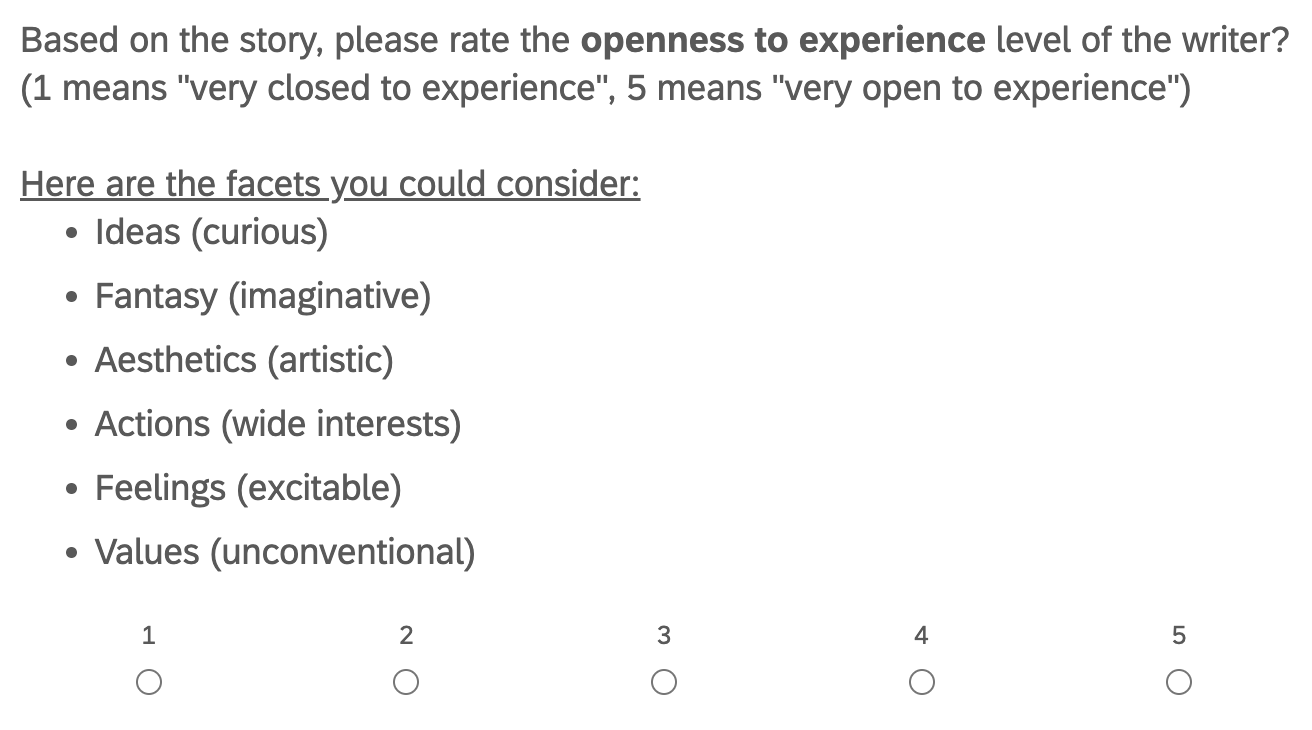}
    \label{fig:openness}
    \caption{Openness question on Prolific.}
\end{figure}

\subsection{Inter-annotator Agreement}
The task of evaluation presents a subjective and complex challenge, which has resulted in a low inter-annotator agreement (IAA) in Krippendorff’s $\alpha$ among the five annotators. We have included the IAA scores for six distinct metrics in Table \ref{tab:iaa_human_six_metrics}. Additionally, the IAA scores for five personality traits are presented in Table \ref{tab:iaa_human_five_trait}. 
The phenomenon of low inter-rater agreeability is consistent with previous findings in labeling tasks for social computing, calling for more attention to creating techniques to navigate the annotation disagreements in order to ensure more accurate label representation \cite{gordon2021disagreement}.

\subsection{Annotator Demographics}
\label{appendix:demographics}

\begin{figure}[h]
    \centering
    \begin{subfigure}[b]{0.3\textwidth}
        \includegraphics[width=\textwidth]{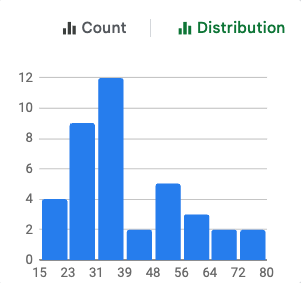}
        \caption{Age}
    \end{subfigure}
    \hfill
    \begin{subfigure}[b]{0.3\textwidth}
        \includegraphics[width=\textwidth]{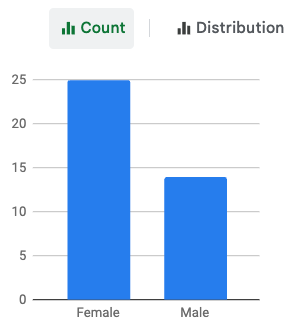}
        \caption{Sex}
    \end{subfigure}
    \hfill
    \begin{subfigure}[b]{0.3\textwidth}
        \includegraphics[width=\textwidth]{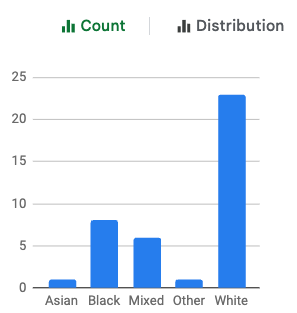}
        \caption{Ethnicity}
    \end{subfigure}
    \caption{Distribution of age, sex, and ethnicity among the 39 Prolific annotators who evaluate the stories.}
    \label{fig:demographics}
\end{figure}

We also include the demographics of 39 unique participants who contribute to evaluate the stories. All of these participants are living in the United States and 37 out of 39 were born in the USA and 2 out of 39 born in Nigeria. We include the distribution of age, sex, and ethnicity in Figure \ref{fig:demographics}.

\section{LLM as Evaluators}
\label{appendix:llm_evaluator_check}

\begin{table*}[ht!]
    \centering
    \small
    \begin{tabular}{ccccccc}
        {\textbf{Evaluator}}  & 
        {\textbf{Readability}}  &
        {\textbf{Redundancy}} & 
        {\textbf{Cohesiveness}} & 
        {\textbf{Likability}} &
        {\textbf{Believability}} & 
        {\textbf{Personalness}} \\
        \Xhline{3\arrayrulewidth}
        \multicolumn{7}{c}{\textit{Evaluation Scores (Mean\textsubscript{STD})}} \\ 
        GPT-3.5 (T=0.0) & $4.75_{0.43}$ & $3.04_{0.40}$ & $4.97_{0.17}$ & $4.22_{0.48}$ & $3.93_{0.25}$ & $3.55_{0.61}$\\
        GPT-3.5 (T=0.3) & $4.70_{0.46}$ & $3.07_{0.54}$ & $4.96_{0.19}$ & $4.26_{0.50}$ & $3.93_{0.30}$ & $3.51_{0.65}$\\
        GPT-3.5 (T=0.7) & $4.65_{0.49}$ & $3.04_{0.63}$ & $4.91_{0.28}$ & $4.29_{0.51}$ & $3.90_{0.41}$ & $3.38_{0.73}$\\
        GPT-3.5 (T=1.0) & $4.54_{0.52}$ & $3.02_{0.85}$ & $4.86_{0.35}$ & $4.27_{0.56}$ & $4.01_{0.43}$ & $3.47_{0.75}$\\
        GPT-4 (T=0.0) & $4.94_{0.24}$ & $4.96_{0.22}$ & $5.00_{0.00}$ & $4.84_{0.36}$ & $4.93_{0.25}$ & $5.00_{0.00}$\\
        GPT-4 (T=0.3) & $4.93_{0.25}$ & $4.95_{0.25}$ & $5.00_{0.00}$ & $4.82_{0.41}$ & $4.94_{0.24}$ & $4.99_{0.08}$\\
        GPT-4 (T=0.7) & $4.87_{0.34}$ & $4.91_{0.33}$ & $5.00_{0.00}$ & $4.78_{0.46}$ & $4.93_{0.25}$ & $4.98_{0.14}$\\
        GPT-4 (T=1.0) & $4.82_{0.38}$ & $4.86_{0.45}$ & $5.00_{0.00}$ & $4.78_{0.43}$ & $4.86_{0.35}$ & $4.98_{0.14}$\\

        
    \end{tabular}
    \caption{LLM evaluation results of GPT-4 generated personal stories with different temperatures. For each evaluated attribute, we report its mean Likert scale and the standard deviation.}
    \label{tab:llm_evaluator_scores_temperature}
\end{table*}

\subsection{Temperature}
We experiment with different temperatures with the GPT-3.5 and GPT-4 evaluators and observe similar trends reported by \citet{chiang2023can}. As shown in Table \ref{tab:llm_evaluator_scores_temperature}, we notice that the ratings given by LLM evaluators are negatively correlated to the temperature. Larger temperature also leads to large variance in the ratings among three LLM evaluators. We set the temperature to 0 in our experiment to ensure the results are more deterministic and reproducible for future research.









\section{LLaMA 2 Results in BFI Scores}
\label{appendix:bfi_llama2}
In this section, we provide additional results of LLaMA 2's performance in BFI assessment in Table \ref{tab:llama2_bfi_assessment}. Overall, LLaMA 2 Persona's BFI assessment shows less score divergence for each trait pair. Even though statistical significance is found for all personality dimensions, their effect size is much smaller when compared with GPT results. 

\begin{table*}[ht]
    \centering
    \small
    \begin{tabular}{ccccccc}
        {\textbf{Persoanlity Trait}}  & 
        {\textbf{Extraversion}}  &
        {\textbf{Agreeableness}} & 
        {\textbf{Conscientiousness }} & 
        {\textbf{Neuroticism}} &
        {\textbf{Openness to Experience}} \\

        \Xhline{3\arrayrulewidth}
        High & $4.55_{0.36}$ & $4.14_{0.21}$ & $4.05_{0.22}$ & $2.70_{0.46}$ & $4.47_{0.24}$ \\
        Low & $2.84_{0.76}$ & $3.69_{0.43}$ & $3.73_{0.33}$ & $2.06_{0.32}$ & $3.58_{0.42}$ \\ \hline
        Cohen's $d$ & $2.86$ & $1.34$ & $1.16$ & $1.63$ & $2.61$ \\ \hline
        
    \end{tabular}
    \caption{We report the statistics of LLaMA 2 Personas' BFI assessment in this Table. The high and low represent the binary traits for each personality dimension. For instance, ``High'' and ``Low'' in Extraversion mean extroverted and introverted, respectively. The effect size of the differences between the two traits is also reported.}
    \label{tab:llama2_bfi_assessment}
\end{table*}

\section{Additional Results in Personality Traits}
\label{appendix:bfi_scores_liwc}
In addition to reporting the significant LIWC features correlated with the binary label in the main paper, we conduct a similar analysis between the LIWC features and the LLM personas 5-point BFI results with Spearsman's $\rho$ and report the findings here.

\subsection{GPT-3.5 Personas}

\noindent\textbf{Extroversion}\space\space Extroverted LLM personas tend to exhibit more social and prosocial behavior in their writings (social: $\rho = 0.27, p < .001$; prosocial: $\rho = 0.18, p < .005$). Introverted personas tend to use features that show authenticity, such as words that are genuine (authentic: $\rho = -0.40, p < .001$). Further, extroverted personas use positive tone and affect more in their writings (affect: $\rho = 0.46, p < .001$; tone\_pos: $\rho = 0.33, p < .001$).

\noindent\textbf{Agreeableness}\space\space Agreeable personas show a strong positive affect and tone in writings (emo\_neg:  $\rho = -0.66, p < .001$; tone\_pos:  $\rho = 0.50, p < .001$). More, they tend to have less conflict-related words (conflict:  $\rho = -0.66, p < .001$), such as fight, and have less differentiation in sentences (differ: $\rho = -0.39, p < .001$), such as ``but'' or ``no''. They also have more prosocial word uses (prosocial: $\rho = 0.34, p < .001$), however, less authenticity (authentic: $\rho = -0.24, p < .001$).

\noindent\textbf{Conscientiousness}\space\space Unconscientious personas have more negative tone and emotion in their writings, such as anger (tone\_neg: $\rho = -0.40, p < .001$; emo\_neg: $\rho = -0.39, p < .001$; emo\_anger: $\rho =-0.43, p < .001$). Their writings tend to use more words that reflect conflicts (conflict: $\rho = -0.41, p < .001$). Conscientious personas use less negation words (negate: $\rho = -0.26, p < .001$), such as ``no'', and have less power related words (power: $\rho = -0.24, p < .001$), such as ``own'' and ``order''. Moreover, conscientious personas exhibit more analytical thinking in the writings (analytic: $\rho = 0.22, p < .001$).

\noindent\textbf{Neuroticism}\space\space The strongest correlated linguistic features for neurotic personas is mental health related words, such as trauma or depression (mental: $\rho = 0.46, p < .001$). Overall, neurotic personas tend to have a negative emotion and tone in their writings (emo\_neg: $\rho = 0.26, p < .001$; tone\_neg: $\rho = 0.22, p < .001$). They also tend to use more words to suggest tentative actions, such as ``if'' or ``any'' (tenta: $\rho = 0.18, p < .005$). Emotionally stable personas are more likely to use words that are related to memory functions, such as ``remember'' (memory: $\rho = -0.15, p < .01$).

\noindent\textbf{Openness}\space\space Open-minded personas tend to have more curiosity driven actions in their writing (curiosity: $\rho = 0.28, p < .001$), such as ``seek'', and more positive tones. Their writings have less conflict-driven words and more affiliation drives (conflict: $\rho = -0.17, p < .005$; affiliation: $\rho = 0.16, p < .005$). Further, open-minded personas tend to write about leisure activities (leisure: $\rho = 0.21, p < .001$), such as ``game'' and ``play''.

\subsection{GPT-4 Personas}

\noindent\textbf{Extroversion}\space\space Introverted personas have more descriptions of their perception in the writings, for instance, their auditory experience (space: $\rho = -0.38, p < .001$; perception: $\rho = -0.38, p < .001$; auditory: $\rho = -0.39, p < .001$). Extroverted personas wrote more future focused event, such as the usage of ``going to'' (focusfuture: $\rho = 0.36, p < .001$). On the usage of pronouns, extroverted personas use more ``we'' while introverted personas tend to use ``I''. Extroverted personas also have more positive tones (tone\_pos: $\rho = 0.21, p < .001$), and use words that are related to rewards or achievement more frequently (reward: $\rho = 0.26, p < .001$; achieve: $\rho = 0.25, p < .001$). 

\noindent\textbf{Agreeableness}\space\space Agreeable personas display more positive tone and emotion in the writings (tone\_pos: $\rho = 0.46, p < .001$; emo\_pos: $\rho = 0.42, p < .001$). They are more prosocial (prosocial: $\rho = 0.29, p < .001$), and use less words that suggest conflict and  more words that show affiliation (conflict: $\rho = -0.51, p < .001$; affiliation: $\rho = 0.22, p < .001$; differ: $\rho = -0.26, p < .001$). Antagonistic personas uses more words that suggest power and ownership, such as ``own'' and ``order''.

\noindent\textbf{Conscientiousness}\space\space Conscientious personas have more prosocial and less negative linguistic features in their writings (prosocial: $\rho = 0.28, p < .001$; tone\_neg: $\rho = -0.34, p < .001$). The writings have less perceived genuineness (authentic: $\rho = -0.24, p < .001$). Further, the writings involve achievement and work related words more frequently (achieve: $\rho = 0.36, p < .001$; work: $\rho = 0.32, p < .001$; reward: $\rho = 0.25, p < .001$).

\noindent\textbf{Neuroticism}\space\space Neurotic personas writings reflect more negative emotions and tones, such as anxiety (emo\_neg: $\rho = -0.59, p < .001$; tone\_neg: $\rho = -0.57, p < .001$; emo\_anx: $\rho = -0.53, p < .001$). The writings have more frequent usage of ``I'' and less usage of ``we'' (i: $\rho = 0.36, p < .001$; we: $\rho = -0.28, p < .001$). Emotionally stable personas write with more prosocial and social behaviors (prosocial: $\rho = -0.27, p < .001$; social: $\rho = -0.28, p < .001$), and the writings have a higher score for perceived genuineness (authentic: $\rho = 0.28, p < .001$). 

\noindent\textbf{Openness}\space\space Open-minded persona's writings have more curiosity and allure-driven linguistics, such as ``research'' and ``wonder''(curiosity: $\rho = 0.55, p < .001$; allure: $\rho = -0.30, p < .001$). Further, the writings contain more analytical thinking and sharing thoughts (analytical: $\rho = 0.27, p < .001$; insight: $\rho = 0.25, p < .001$). Open-minded personas write with more big words with seven letters or longer and more words per sentence (BigWord: $\rho = 0.29, p < .001$; WPS: $\rho = 0.26, p < .001$).

\subsection{LLaMA 2 Personas}
\noindent\textbf{Extroversion}\space\space Extroverted personas' writings have more positive tones (emo\_pos: $\rho = 0.46, p < .001$; tone: $\rho = 0.47, p < .001$). Further, the writings tend to have more social references and affiliations (scorefs: $\rho = 0.48, p < .001$; affiliation: $\rho = 0.48, p < .001$; emo\_pos: $\rho = 0.50, p < .001$; differ: $\rho = -0.36, p < .001$;). There is a weak association with more acclimations, and extroverted personas' writings show more drive-related words (exclam: $\rho = 0.23, p < .001$; Drive: $\rho = 0.34, p < .001$).

\noindent\textbf{Agreeableness}\space\space Agreeable personas have more positive emotion and tone in writings (emo\_pos: $\rho = 0.50, p < .001$; tone: $\rho = 0.41, p < .001$). There is more word usage around friends and less about work (friend: $\rho = 0.47, p < .001$; work: $\rho = -0.48, p < .001$). Similar to the previous LMs, it shows more affiliation and fewer conflicts from the writings (conflict: $\rho = -0.31, p < .001$; affiliation: $\rho = 0.25, p < .001$). There is more usage of pronouns like ``we'' than ``they'' (they: $\rho = -0.27, p < .001$; we: $\rho = 0.20, p < .001$). Interestingly, the writings reflect a slight negative correlation with prosociality (prosocial: $\rho = 0.22, p < .001$).

\noindent\textbf{Conscientiousness}\space\space Similar to the trend in GPT-3.5 and GPT-4, LLaMA 2's personified writings have little significant linguistic features with Conscientiousness. It has a weak correlation in word usage around home and fulfillment (home: $\rho = 0.17, p = .002$; fulfill: $\rho = 0.15, p = .006$). Further, a weak link is found for familiar and friendly words (friend: $\rho = 0.15, p = .007$; family: $\rho = 0.13, p = .02$). 

\noindent\textbf{Neuroticism} \space\space The neurotic personas's writings show a strong negative correlation with positive emotions (emo\_pos: $\rho = -0.47, p < .001$; tone\_neg: $\rho = 0.43, p < .001$). Further, word usage around friend, leisure, and social references is negatively correlated with the Persona's neurotic scores, while work and insight-related words are positively related (friend: $\rho = -0.47, p < .001$; leisure: $\rho = -0.47, p < .001$; socrefs: $\rho = -0.43, p < .001$; work: $\rho = -0.42, p < .001$; insight: $\rho = -0.43, p < .001$).

\noindent\textbf{Openness} \space\space Open-minded LLaMA 2 Personas' writings show a positive trend with positive emotions and affiliations (emo\_pos: $\rho = 0.43, p < .001$; affiliation: $\rho = 0.34, p < .001$). Interestingly, it shows a negative trend of using words related to work or achievement (work: $\rho = -0.54, p < .001$; achieve: $\rho = -0.42, p < .001$). Further, the writings are more likely to reflect curiosity and have more social references (curiosity: $\rho = 0.29, p < .001$; socrefs: $\rho = 0.29, p < .001$).

\end{document}